\documentclass[sigconf]{acmart}
\usepackage{epstopdf}
\usepackage{multicol}
\usepackage{amsmath}
\usepackage{multirow}
\usepackage{balance}
\AtBeginDocument{%
  \providecommand\BibTeX{{%
    \normalfont B\kern-0.5em{\scshape i\kern-0.25em b}\kern-0.8em\TeX}}}

\copyrightyear{2021}
\acmYear{2021}
\setcopyright{acmlicensed}\acmConference[MM '21]{Proceedings of the 29th ACM International Conference on Multimedia}{October 20--24, 2021}{Virtual Event, China}
\acmBooktitle{Proceedings of the 29th ACM International Conference on Multimedia (MM '21), October 20--24, 2021, Virtual Event, China}
\acmPrice{15.00}
\acmDOI{10.1145/3474085.3475601}
\acmISBN{978-1-4503-8651-7/21/10}

\begin{document}
\fancyhead{}
\title{TriTransNet: RGB-D Salient Object Detection with a Triplet Transformer Embedding Network}

\author{Zhengyi Liu}
\authornote{Corresponding author.}

\orcid{}
\affiliation{%
  \institution{School of Computer Science and Technology,
Anhui University}
  \streetaddress{}
  \city{Hefei}
  \state{}
  \country{China}
  \postcode{}
}
\email{liuzywen@ahu.edu.cn}

\author{Yuan Wang}

\affiliation{%
  \institution{School of Computer Science and Technology,
Anhui University}
  \streetaddress{}
  \city{Hefei}
  \country{China}}
\email{wangyuan.ahu@qq.com}

\author{Zhengzheng Tu}
\affiliation{%
  \institution{School of Computer Science and Technology,
Anhui University }
  \city{Hefei}
  \country{China}
}
\email{15352718@qq.com}

\author{Yun Xiao}
\affiliation{%
 \institution{School of Computer Science and Technology,
Anhui University}
 \streetaddress{}
 \city{Hefei}
 \state{}
 \country{China}}
\email{280240406@qq.com}

\author{Bin Tang}
\affiliation{%
 \institution{School of Artificial Intelligence and Big Data, Hefei University}
 \streetaddress{}
 \city{Hefei}
 \state{}
 \country{China}}
\email{424539820@qq.com}
\renewcommand{\shortauthors}{Zhengyi Liu, et al.}

\begin{abstract}
Salient object detection is the pixel-level dense prediction task which can highlight the prominent object in the scene. Recently U-Net framework is widely used, and continuous convolution and pooling operations generate multi-level features which are complementary with each other. In view of the more contribution of high-level features for the performance, we propose a triplet transformer embedding module to enhance them by learning long-range dependencies across layers. It is the first to use three  transformer encoders with shared weights to enhance multi-level features. By further designing scale adjustment module to process the input, devising three-stream decoder to process the output and attaching depth features to color features for the multi-modal fusion, the proposed triplet transformer embedding network (TriTransNet) achieves the state-of-the-art performance in RGB-D salient object detection, and pushes the performance to a new level. Experimental results demonstrate the effectiveness of the proposed modules and the competition of TriTransNet.\footnote{The code is available at https://github.com/liuzywen/TriTransNet.}
\end{abstract}



\maketitle

\section{Introduction}
Salient object detection (SOD) simulates the visual attention mechanism to capture the prominent object in the scene. It has been widely applied in the computer vision tasks, such as image segmentation~\cite{donoser2009saliency}, tracking~\cite{ma2017saliency,hong2015online,zhang2020non}, retrieval~\cite{gao2015database}, compression~\cite{ji2013video}, edit~\cite{wang2018deep} and quality assessment~\cite{jiang2017optimizing}.

\begin{figure}[!htp]
\centering
\begin{tabular}{c}
\includegraphics[width =0.8\linewidth]{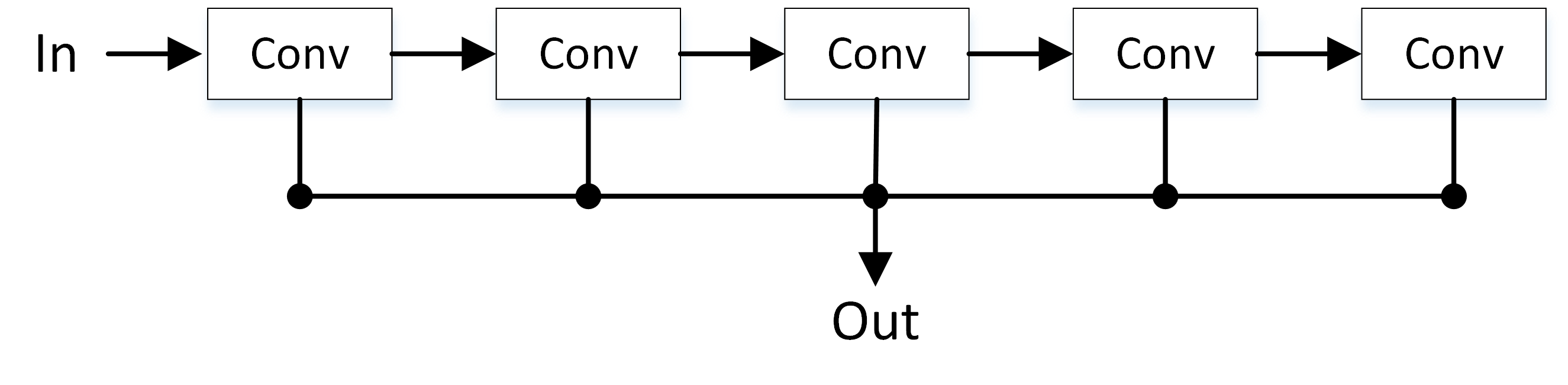}\\
(a) U-Net framework\\\\
\includegraphics[width = 0.8\linewidth]{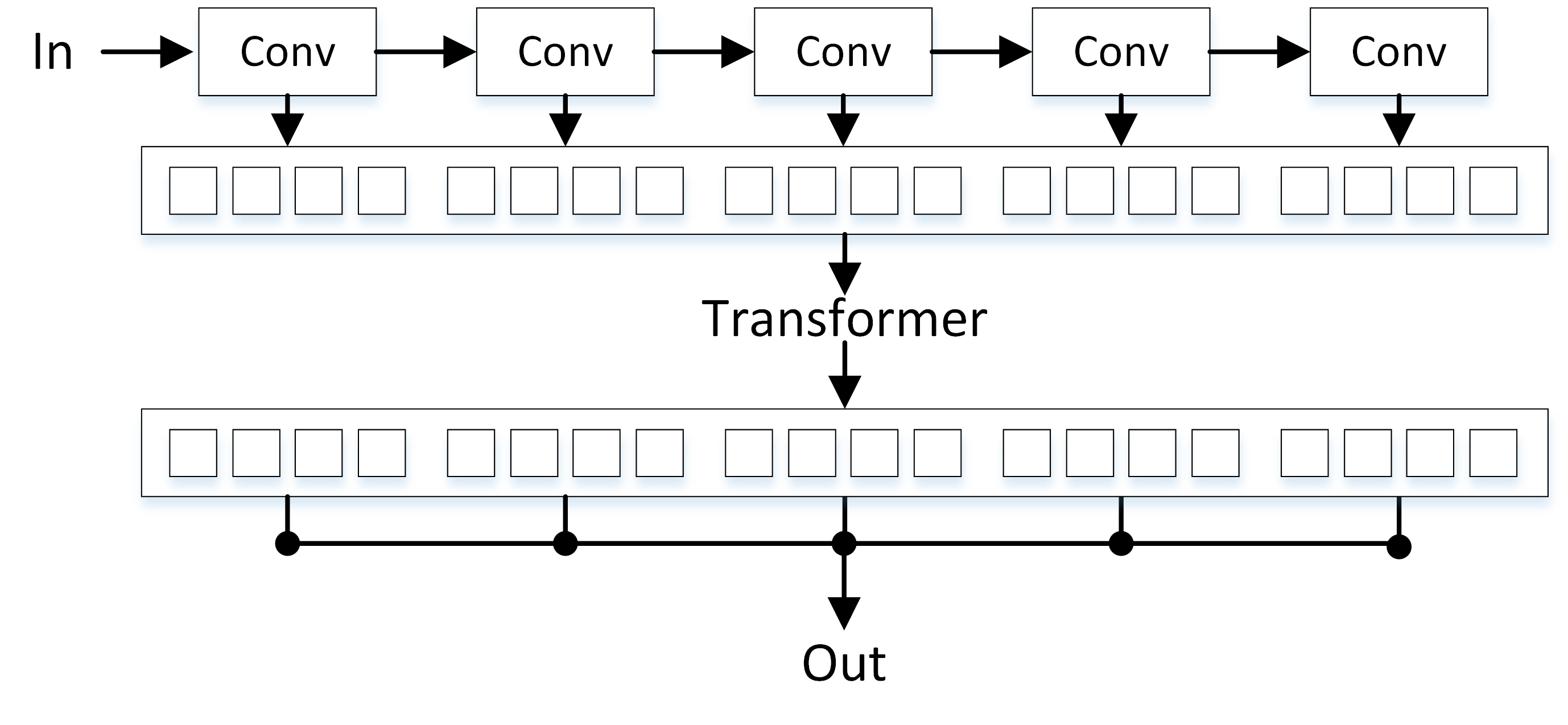}\\
(b) visual-transformer-FPN (VT-FPN)\\\\
\includegraphics[width = 0.8\linewidth]{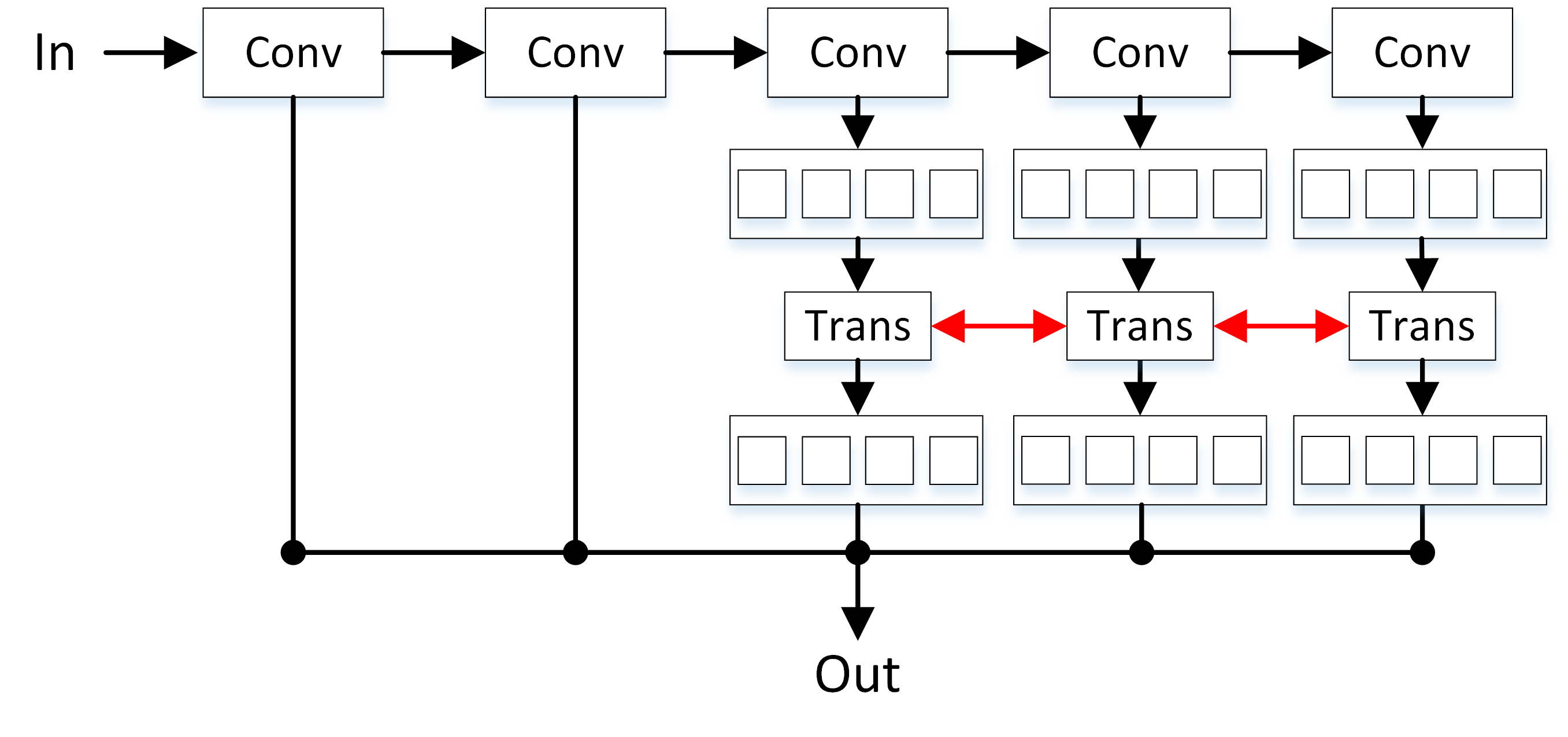}\\
(c) Our proposed  triplet transformer embedding network
\end{tabular}
\caption{Comparison between U-Net framework, VT-FPN and our proposed network.}
\label{fig:FrameCompare}
\end{figure}

As a pixel-level dense prediction task, salient object detection usually uses CNN based U-Net framework\cite{ronneberger2015u} (Fig.~\ref{fig:FrameCompare}(a)) to encode images from low-level to high-level, and then decode back to the full spatial resolution.
Research\cite{wu2019cascaded} points out that the performance tends to saturate quickly when gradually aggregating features from high-level to low-level. In other words, high-level features contribute more to the performance. 
Therefore, we propose a triplet transformer embedding module (TTEM) to enhance the feature representation of high three layers.

As we all known,
Transformer\cite{NIPS2017_3f5ee243} has recently attracted a lot of attention in computer vision domain, but it is  also encountering  high computational cost problem. PVT\cite{wang2021pyramid} adopts a spatial-reduction attention (SRA) layer to reduce the resource cost to learn high-resolution feature maps. CvT\cite{wu2021cvt} introduces convolutional into the Vision Transformer architecture to concurrently maintain a high degree of computational and memory efficiency.
Swin Transformer\cite{liu2021swin} uses the shifted windows calculation method to propose a hierarchical Transformer, which has the flexibility of modelling at various scales and has linear computational complexity relative to the image size.
Multi-Scale Vision Longformer\cite{zhang2021multi} proposes multi-scale coding structure, and further improves its attention mechanism to reduce the computational and memory cost.

Unlike these profound designs, we introduce Transformer into U-Net framework to enhance the features of high three layers, which can be easily integrated into existing U-Net framework for significant improvement with less cost.
The features of  high three layers show the different attributions but the same in nature, which are the different aspects of the same input image. The proposed triplet transformer embedding module (TTEM) is composed of three  standard transformer encoders\cite{NIPS2017_3f5ee243} with shared weights.  It is beneficial to find the common information which is hidden in the multi-level features and achieve the better fusion by learning long-range dependencies across levels.

Taking TTME as the core, we further propose the triplet transformer embedding network (TriTransNet). At first, multi-level features are adjusted to the same size by a transition layer and progressively upsampling fusion module. Second, features are fed into TTME to be enhanced. Last, the output features of TTME and the features of low two layers are effectively fused by a three-stream decoder.

Our proposed  TriTransNet is  the first attempt to use three standard transformer encoders with shared weights to enhance the feature representation.
Different from visual-transformer-FPN (VT-FPN)\cite{wu2020visual} in Fig.~\ref{fig:FrameCompare}(b) which merges visual tokens from feature map of each layer with one transformer, our TriTransNet in Fig.~\ref{fig:FrameCompare}(c) adopts weight sharing strategy to make visual tokens extracted from multi-level features more abundant enough to express the original information, and meanwhile high-layer semantic information and middle-layer texture or shape information are both better excavated by parallel self-attention mechanism.

In addition, depth information is proved to supply the useful cues and boost the performance for saliency detection~\cite{ouerhani2000computing}, especially in some challenging
and complex scenarios, e.g. the low color contrasts between salient
objects and background, the cluttered background interferences.
But depth image with poor quality, which likes a noise, brings some negative influences~\cite{fan2020rethinking}.
Following depth guided manners~\cite{piao2020a2dele,chen2021global,sun2020real,pan2020cross,fan2020bbs,chen2020progressively,zhao2019contrast,zeng2019deep,zhu2019pdnet}
we design depth purification module, which uses depth information to purify the color features.

Our main contributions can be summarized as follows:
\begin{itemize}
  \item
  A triplet transformer embedding module is proposed and embedded into CNN based U-Net framework to enhance the feature representation. It is composed of three standard transformer encoders with shared weights, learning the common information from multi-level features.
  \item
  Based on the proposed triplet transformer embedding module, triplet transformer embedding network is designed to detect the salient objects in RGB-D image. Multi-level features from encoder need to be adjusted to the same size by a transition layer and progressively upsampling fusion module, and then fed into triplet transformer embedding module. Then the output of triplet transformer embedding module need to be combined with the features of low two layers by three-stream decoder to achieve the decoding process.
  \item
  Depth image is viewed as the supplement to color feature, and attached to color feature to enhance the feature representation by depth purification module which introduces spatial attention and channel attention.
  \item Due to the advantage of the proposed triplet transformer embedding module, the proposed model pushes the performance of RGB-D salient object detection to a new level and shows the state-of-the-art performance on several public datasets.
\end{itemize}

\section{Related work}\label{section2}
\subsection{RGB-D saliency detection}
In RGB-D image, color image provides appearance and texture information, and depth image contains 3D layout and spatial structure.
The fusion of color feature and depth feature is always an important issue in RGB-D saliency detection.
References\cite{liu2019salient, zhao2020single, zhang2020bilateral, chen2021rd3d} use early fusion or input fusion,
references\cite{chen2018progressively,chen2020progressively,li2020icnet,li2020rgb,li2020asif}
employ two-stream subnetwork to achieve the middle fusion,
references\cite{zhu2019pdnet,zhao2019contrast,piao2020a2dele,wu2020mobilesal,pang2020hierarchical}
apply depth guided fusion and references~\cite{wang2019adaptive,liu2020cross,chen2020dpanet} adopt late fusion.

Although depth information can supply the useful cues  for saliency detection~\cite{ouerhani2000computing}, depth image with poor quality can bring some negative influences too~\cite{fan2020rethinking}.
In order to solve the filtering issue of low-quality depth map,
D3Net~\cite{fan2020rethinking} uses gate mechanism to filter the poor depth map,
EF-Net~\cite{chen2020ef} enhances the depth maps by color hint map,
DQSD~\cite{chen2020depth}  integrates a depth quality aware subnetwork into the classic bi-stream structure, assigning the weight of depth feature  before conducting the fusion.
In addition, CoNet\cite{ji2020accurate},  DASNet~\cite{zhao2020depth}, SSDP\cite{wang2020synergistic}  and MobileSal~\cite{wu2020mobilesal} introduce depth estimation, learning to detect the salient object simultaneously.

In the paper, we adopt depth guided manner. Depth information is viewed as the supplement to the color feature. It enhances the color feature by attention mechanism.

\subsection{Transformer}
Transformer is first proposed by\cite{NIPS2017_3f5ee243} to replace recurrent neural networks (RNN), e.g.long short-term memory (LSTM) and gated recurrent unit(GRU) for machine translation tasks. It can overcome intrinsic shortages of RNN and has dominated nature language processing (NLP) field and are becoming
increasingly popular in computer vision tasks, e.g. image classification\cite{dosovitskiy2021an}, object detection\cite{carion2020end}, semantic segmentation\cite{SETR}, line segment\cite{xu2021line}, person re-identification\cite{zhu2021aaformer},
action detection\cite{zhao2021tuber},
image completion\cite{zheng2021tfill}, 3D point cloud processing\cite{zhao2020point,guo2020pct}, pose estimation\cite{stoffl2021end}, facial expression recognition\cite{ma2021robust}, object tracking\cite{meinhardt2021trackformer} etc.
DETR\cite{carion2020end} takes the lead in applying Transformer to the field of object detection and achieves the better performance. The successful use of ViT\cite{dosovitskiy2021an} in image classification tasks has made the research on visual Transformer a hot topics. SETR\cite{zheng2021rethinking} deploys a pure Transformer as the encoder, combined with a simple decoder, to achieve a powerful semantic segmentation model. Besides, TransUNet\cite{chen2021transunet} uses the pre-trained ViT\cite{dosovitskiy2021an} as a powerful backbone of the U-Net\cite{ronneberger2015u} network structure, and performs well in the field of medical image segmentation.

However, pure transformer has great limitations. As a result, many improved visual transformers have emerged. The Conditional Position encodings Visual Transformer (CPVT)\cite{chu2021we} replaces the fixed position encoding in ViT\cite{dosovitskiy2021an} with the proposed conditional position encoding (CPE), which makes it possible for Transformer to process inputs of arbitrary sizes. Tokens-to-Token (T2T)\cite{yuan2021tokens} adopts a novel progressive tetanization mechanism, which models local structural information by aggregating adjacent tokens into one token, while reducing the length of the token. LocalViT\cite{li2021localvit} adds locality to vision transformers by introducing depth-wise convolution into the feed-forward network,
improving a locality mechanism for information exchange within a local region.
Considering that most visual Transformers ignore the inherent structural information inside the sequence of patches, Transformer-iN-Transformer (TNT)\cite{han2021transformer} proposes to use outer Transformer block and inner Transformer block to model patch-level and pixel-level representations, respectively.
Co-Scale Conv-Attentional Image Transformers\cite{xu2021co}  designs a conv-attention module to realize relative position embedding and  enhance computation efficiency, and further proposes a co-scale mechanism to introduce cross-scale attention to enrich multi-scale feature.

On the other hand, CNN has  the advantages of extracting low level features and strengthening locality, while Transformer has the advantages in establishing long-range dependencies. Some research makes full use of both advantages.
TransFuse\cite{zhang2021transfuse} uses a dual-branch structure, which uses Transformer to capture global dependencies, while low-level spatial details are extracted by CNN branches. Similarly, CoTr\cite{xie2021cotr} uses the CNN backbone to extract feature representations and proposes to use deformable Transformer (DeTrans) to model long-range dependencies, effectively bridging the convolutional neural network and Transformer.
ICT\cite{wan2021high} uses transformer to recover pluralistic coherent structures together with some coarse textures, and uses CNN to enhances local texture details of coarse priors, so as to achieve excellent results on the image completion task.
TransT\cite{chen2021transformer} uses Siamese-based CNN network for feature extraction, and designs the self-attention-based ego-context augment (ECA) and cross-attention-based cross-feature augment (CFA) modules for feature fusion.
Compact Transformers\cite{hassani2021escaping} eliminates the requirement for class token and position embedding through a novel sequence pooling strategy and the use of convolutions, so as to perform head-to-head with state-of-the-art CNNs on small datasets.

Follow this strategy, we present triplet transformer embedding module which is embedded into a U-Net framework to improve the performance of RGB-D saliency detection. Combining both advantages, our model achieves the state-of-the-art performance.
\section{Proposed method}
\subsection{Overview}
\begin{figure*}[!htp]
  \includegraphics[width=\textwidth]{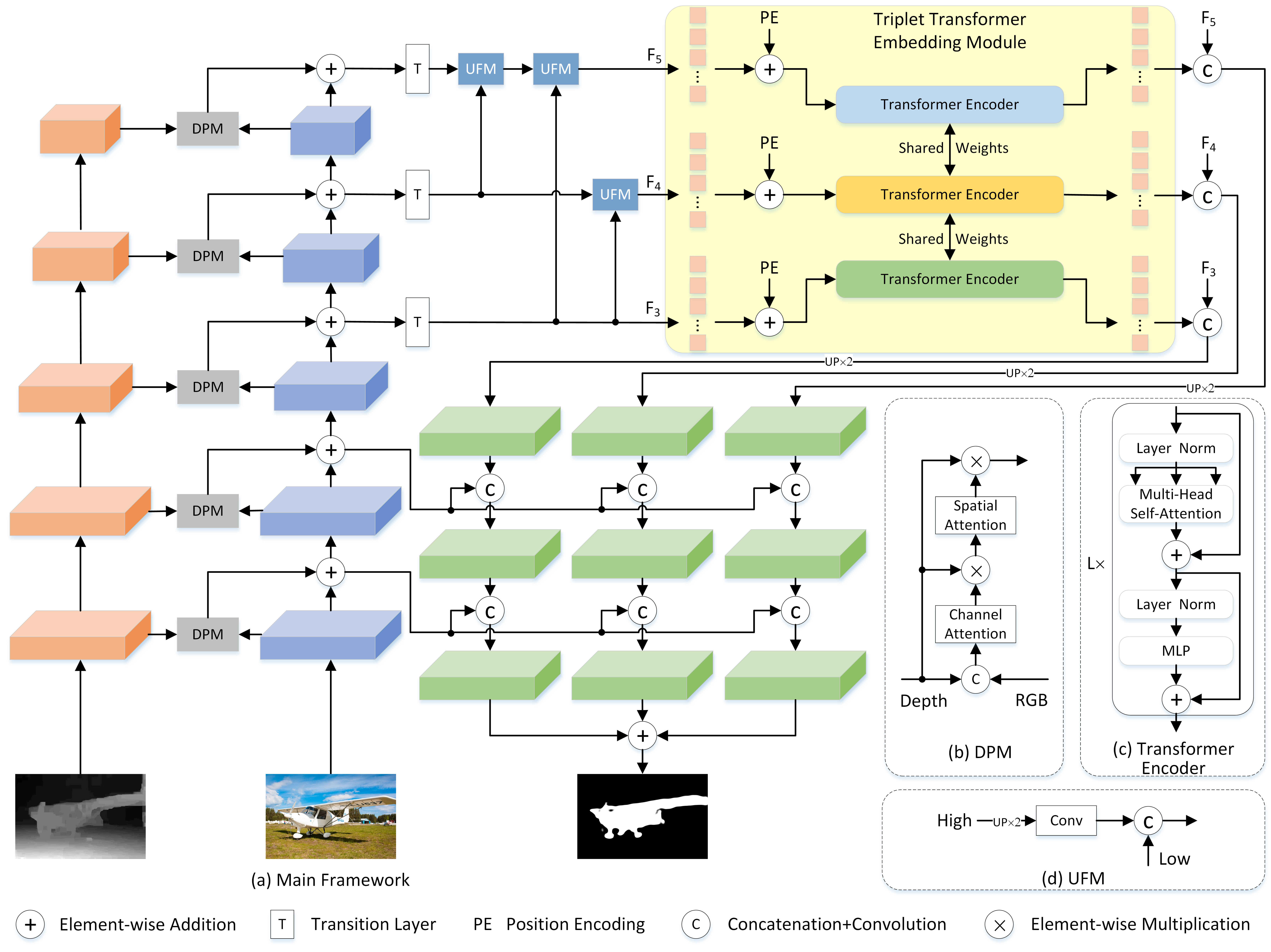}
  \caption{Our proposed triplet transformer embedding network for RGB-D salient object detection.}
  \label{fig:main}
\end{figure*}

The overall framework of the proposed triplet transformer embedding network is depicted in Fig.\ref{fig:main}(a), which consists of multi-modal fusion encoder, feature enhancement module and three-stream  decoder. The details can be seen in the following sections.

\subsection{Multi-modal fusion encoder}
Color and depth image in RGB-D image are two expressions for different modalities of the same scene. Color image provides appearance cue and depth image shows three dimension spatial information. Due to existence of poor quality depth map induced by the imaging devices or conditions, we propose multi-modal fusion encoder, in which depth features are first purified by multi-modal features using attention mechanism, and then served as supplement to the color feature by the residual connection\cite{he2016deep}.
Residual part is designed as depth purification  module
(DPM), and shortcut connection part is used to preserve more original color information.

In DPM which is shown in Fig.~\ref{fig:main}(b), depth feature is concatenated with color feature, and fed into a channel attention module to get attentive channel mask, which is used to purify the depth feature in a channel manner. Next, purified depth feature is fed into a spatial attention module again to generate attentive spatial mask, which is  used to purify the depth feature in a spatial manner. The process can be described as:
\begin{equation}
\begin{aligned}
F^{r}_{i}=f^{d}_{i}\times  SA(f^{d}_{i}\times CA(Cat(f^{d}_{i},f^{r}_{i})))+f^{r}_{i}
\end{aligned}
\end{equation}
where $f^r_i$ and $f^d_i$  represent color and depth features extracted by backbone network respectively in which $i=1,\cdots,5$, $Cat(\cdot)$ denotes concatenation and following convolution operation, $CA(\cdot)$ and $SA(\cdot)$ are channel and spatial attention operation which is proposed by CBAM\cite{woo2018cbam}, ``$\times$" is element-wise multiplication operation, ``$+$" is element-wise addition operation.

Thus, the depth feature with poor quality can be purified, and then attached to color feature to generate more accuracy feature representation $F^{r}_{i}(i=1,\cdots,5)$.
\subsection{Feature enhancement module}
In this module, we first adjust the features of  high three layers to the same size, and then use the triplet transformer embedding module to enhance the feature representation by learning long-range dependency across levels, and last concatenate the input and output of triplet transformer embedding module to preserve more original information.
\subsubsection{Scale adjustment module}
The triplet transformer embedding module is composed of three standard transformer encoders with shared weights. Its input should be the features with the same size.
But the sizes of the multi-level features $F^{r}_{i}$ from multi-modal fusion encoder are the different.
Therefore,  the first important task is to adjust the sizes of multi-level features.

At first, a transition layer which contains a 3$\times$3 convolution and a ReLU activation function is applied on $F^{r}_i$. It can adjust the  number of channels of multi-level features to the same size. It can be described as:
\begin{equation}
\begin{aligned}
F'^{r}_{i}=\sigma(Conv(F^{r}_{i}))\quad i=3,\cdots,5
\end{aligned}
\end{equation}
where $Conv(\cdot)$ is 3$\times$3 convolution operation, and $\sigma(\cdot)$ is ReLU activation function.

Then, we design a progressively upsampling fusion module which is used to adjust the resolution of the  features in the high three layers to the same size.
Since the direct upsampling with 2$\times$ or 4$\times$ ratio will bring some noises,
the features are progressively upsampled and fused.  The fusion process can be described as:
\begin{equation}
\begin{aligned}
&F_5=UFM(UFM(F'^{r}_{5},F'^{r}_{4}),F'^{r}_{3})&\\
&F_4=UFM(F'^{r}_{4},F'^{r}_{3})&\\
&F_3=F'^{r}_{3}&\\
\end{aligned}
\end{equation}
where $UFM(\cdot)$ is shown in Fig.\ref{fig:main}(d). The detail can be described as:\\
\begin{equation}
\begin{aligned}
UFM(F_{high},F_{low})=Cat(Conv(Up(F_{high})), F_{low})\\
\end{aligned}
\end{equation}
where $F_{high}$ and $F_{low}$ denote the feature from the higher layer with low resolution and the feature from the lower layer with high resolution, respectively, and $Up(\cdot)$ denotes 2$\times$upsampling operation.

Compared with direct 2$\times$, 4$\times$upsampling on $F'^{r}_{4}$ and $F'^{r}_{5}$, progressively upsampling fusion module can not only adjust the features to the same resolution but also increase the spatial detail of feature in the high layer by progressive fusion process.

Thus, the features $F_i(i=3,\cdots,5)$ with the same scales will be served as the input and fed into next triplet transformer embedding module.
\subsubsection{Triplet Transformer Embedding Module (TTEM)}
The features are first converted into the sequences of feature embedding, and then fed to three standard transformer encoders with shared weights to model the long-range relationship among different levels, and last reshaped to the original size of features.

Specifically, each input feature $F_i(i=3,\cdots,5)$ are first flattened into a 1D sequence $\{F^p_i|p=1,\cdots, N\}$, where $N$ is the number of patches.
Each patch $F^p_i$ is then mapped into a latent $D$-dimensional embedding space by a trainable linear projection layer. Furthermore, we learn specific position embedding which are added to the patch embedding to retain positional information. The process can be described as:
\begin{equation}
\begin{aligned}
Z^0_i=[F^1_i+PE^1;F^2_i+PE^2;\cdots,;F^N_i+PE^N]
\end{aligned}
\end{equation}
where $PE=\{PE^p|p=1,\cdots,N\}$ is a 1D learnable positional embedding.

The remaining architecture essentially follows the standard transformer encoder\cite{NIPS2017_3f5ee243} which stacks $L$ transformer layer. It is shown in Fig.\ref{fig:main}(c).
Each transformer layer contains multi-headed self-attention (MSA) and multi-layer perceptron (MLP) sublayer. Layer normalization (LN)\cite{ba2016layer} are inserted before these two sublayers, and the residual connection is performed after these two sublayers.
The process can be described as:
\begin{equation}
\left\{\begin{array}{l}
{Z^{l}_i}^{\prime}=MSA\left(LN\left(Z^{l-1}_i\right)\right)+Z^{l-1}_i \\
Z^{l}_i=MLP\left(LN\left({Z^{l}_i}^{\prime}\right)\right)+{Z^{l}_i}^{\prime}
\end{array} \quad l=1, \cdots L\right.
\end{equation}
where $L$ denotes the number of transformer layers in the standard transformer encoder.
\subsubsection{Feature concatenation module}
The outputs of three weights shared  transformer encoders  $Z^{L}_i(i=3,\cdots,5)$ fuses the information of three layers by Transformer mechanism, so as to enhance the original feature representation. In order to preserve the more original information, we further cascade these outputs with original features to generate the enhanced features of high three layers. The process can be described as:
\begin{equation}
\begin{aligned}
&F'_i=Cat(Z^{L}_i,F_i) & \quad i=3, \cdots 5&\\
\end{aligned}
\end{equation}
\subsection{Three-stream decoder}
After the features of high three layers are enhanced by the proposed triplet transformer embedding module, we will
combine them with  the features of low two layers to achieve the decoding process. There are two decoding methods. One is single-stream decoding and  the other is three-stream decoding.
The single-stream decoding first fuses three output results of feature enhancement module, and then combine it with two  features in the low layers.
The three-stream decoding first combines each output result of feature enhancement module  with two features in the low  layers, and then fuses three-stream results.
We conduct two decoding processes, and find three-stream decoding is better than single-stream decoding. Next, we use formula to show three-stream  decoder as follow:
\begin{equation}
\begin{aligned}
F''_i=Cat(Cat(Up(F'_i),F^2_r),F^1_r)& \quad i=3, \cdots 5\\
\end{aligned}
\end{equation}

The above three features   are performed upsampling, convolution operation and  sigmoid function to generate the saliency maps $S_i(i=1,\cdots,3)$ which are supervised by the ground truth maps.
\begin{equation}
\begin{aligned}
S_{i}=sig(Conv(Up(Conv(Up(F''_i)))))\quad
\end{aligned}
\end{equation}
where $sig(\cdot)$ denotes sigmoid function.

At last, we also fuse all the features above to generate the final saliency map.
\begin{equation}
\begin{aligned}
S_{final}=sig(\sum_{i=3}^{5}{Conv(Up(Conv(Up(F''_i))))})\\
\end{aligned}
\end{equation}

Pixel position aware loss $L^{s}_{ppa}$~\cite{wei2020f3net} is adopted for end-to-end training. The whole loss is defined as:
\begin{equation}
\begin{aligned}
L = L^{s}_{ppa}(S_{final},G)+\sum_{i=3}^{5}{L^{s}_{ppa}(S_i,G)}
\end{aligned}
\end{equation}
where $G$ is ground truth saliency map.
\section{Experiments}

\subsection{Datasets and evaluation metrics}

\subsubsection{Datasets}
We evaluate the proposed method on six challenging RGB-D SOD datasets.
NLPR~\cite{peng2014rgbd} includes  1000 images with single or multiple salient objects.
NJU2K~\cite{ju2014depth} consists of 2003 stereo image pairs and ground-truth maps with different objects, complex and challenging scenes.
STERE~\cite{niu2012leveraging} incorporates 1000 pairs of binocular images downloaded from the Internet.
DES~\cite{cheng2014depth} has 135 indoor images collected by
Microsoft Kinect.
SIP~\cite{fan2020rethinking} contains 1000 high-resolution images of multiple salient persons.
DUT~\cite{piao2019depth} contains 1200 images captured by Lytro camera in real life scenes.

For the sake of fair comparison, we use the same training dataset as in~\cite{fan2020rethinking,chen2020progressively}, which consists of 1,485 images from the NJU2K dataset and 700 images from the NLPR dataset. The remaining images in the NJU2K and NLPR datasets and the whole datasets of STERE, DES and SIP are used for testing.
In addition, on the DUT dataset, we follow
the same protocols as in ~\cite{piao2019depth, zhao2020single,piao2020a2dele,li2020rgb,ji2020accurate} to add additional 800 pairs from DUT for training and test on the
remaining 400 pairs.
In summary, our training set contains 2,185 paired RGB and depth images, but when testing is conducted on DUT, our training set contains 2,985 paired ones.
\subsubsection{Evaluation Metrics}
We adopt five widely used metrics to evaluate the performance of our model and other  state-of-the-art RGB-D SOD models, including the precision-recal(PR) curve~\cite{borji2015salient}, E-measure~\cite{fan2018enhanced}, S-measure~\cite{fan2017structure},
F-measure~\cite{achanta2009frequency} and mean absolute error (MAE)~\cite{perazzi2012saliency}. Specifically, the PR curve plots precision and recall values by  setting a series of thresholds on the saliency maps to get the binary masks and further comparing them with the ground truth maps.
The E-measure simultaneously captures global statistics and local pixel matching information.
The S-measure can evaluate both region-aware and object-aware structural similarity between saliency map and ground truth.
The F-measure is the weighted harmonic mean of precision and recall, which can evaluate the overall performance.
The MAE measures the average of the per-pixel absolute difference between the saliency maps and the ground truth maps. In our experiment, E-measure and F-measure adopts adaptive values.
\subsection{Implementation details}
During the training and testing phase, the input RGB and depth images are resized to 256$\times$256. Multiple enhancement strategies are used for all training images, i.e. random flipping, rotating and border clipping. Parameters of the backbone network are initialized with the pretrained parameters of ResNet-50 network~\cite{he2016deep}. The hyper-parameters in transformer encoder are set as: $L=12$,$D=768$,$N=1024$.
The rest of parameters are initialized to PyTorch default settings. We employ the Adam optimizer~\cite{kingma2014adam} to train our network with a batch size of 3 and an initial learning rate 1e-5, and the learning rate will be divided by 10 every 60 epochs.  Our model is trained on a machine with a single NVIDIA GTX 3090 GPU. The model converges within 150 epochs, which takes nearly 15 hours.
\subsection{Comparisons with the state-of-the-art}
Our model is compared with 16 state-of-the-art RGB-D SOD models, including  D3Net~\cite{fan2020rethinking}, ICNet~\cite{li2020icnet}, DCMF~\cite{chen2020rgbd}, DRLF~\cite{wang2020data}, SSF~\cite{zhang2020select}, SSMA~\cite{liu2020learning}, A2dele~\cite{piao2020a2dele}, UCNet~\cite{zhang2020uc}, CoNet~\cite{ji2020accurate}, DANet~\cite{zhao2020single},
JLDCF\cite{fu2020jl}, EBFSP\cite{huang2021employing},CDNet\cite{jin2021cdnet},
HAINet\cite{li2021hierarchical}, RD3D\cite{chen2021rd3d} and DSA2F\cite{Sun2021DeepRS}. To ensure the fairness of the comparison results, the saliency maps of the evaluation are provided by the authors or generated by running source codes.

\subsubsection{Quantitative Evaluation.} 
\begin{figure*}[!htp]
\centering
\begin{tabular}{ccc}
\includegraphics[width = 0.33\textwidth]{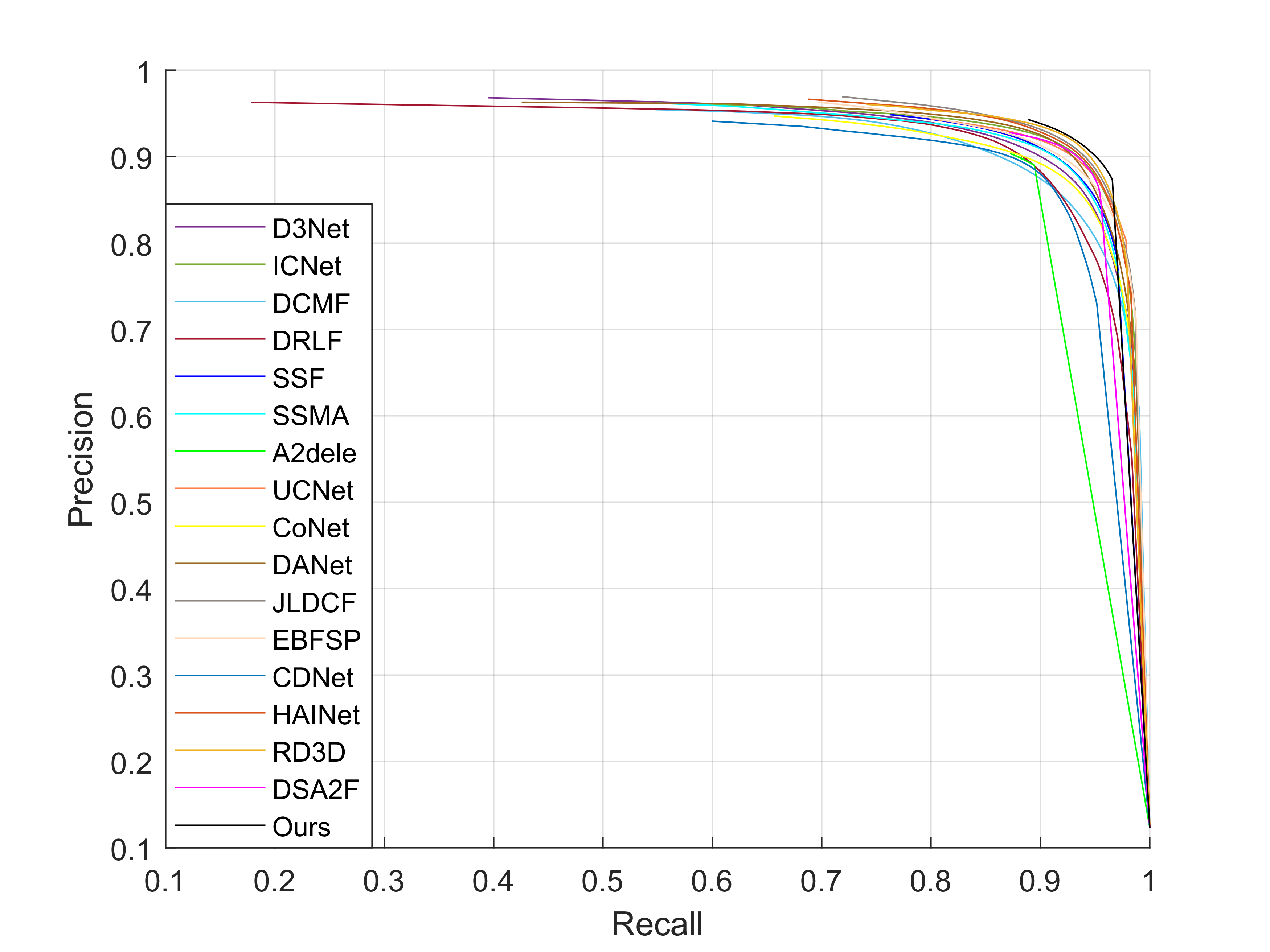}&\includegraphics[width = 0.33\textwidth]{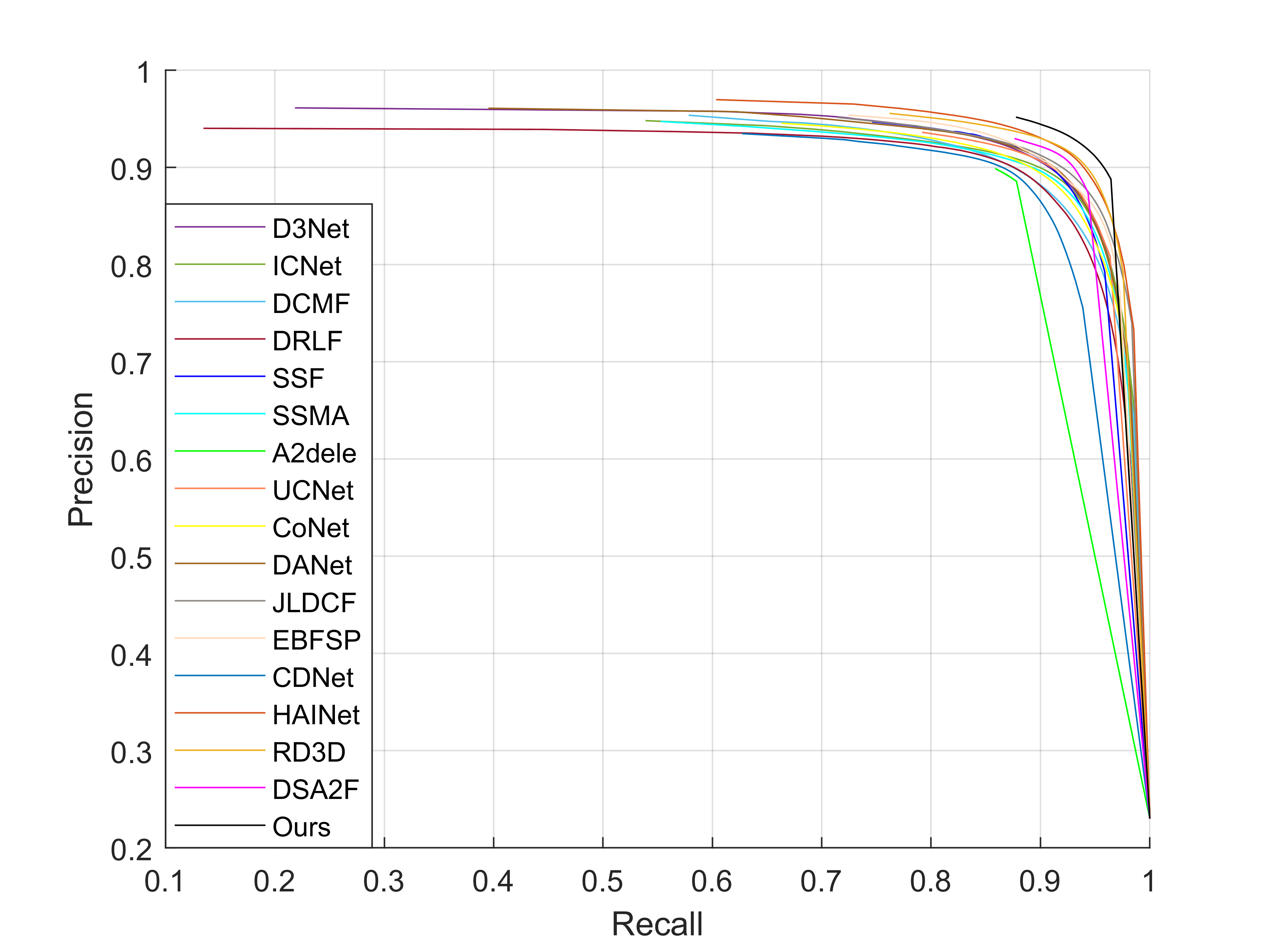}&\includegraphics[width = 0.33\textwidth]{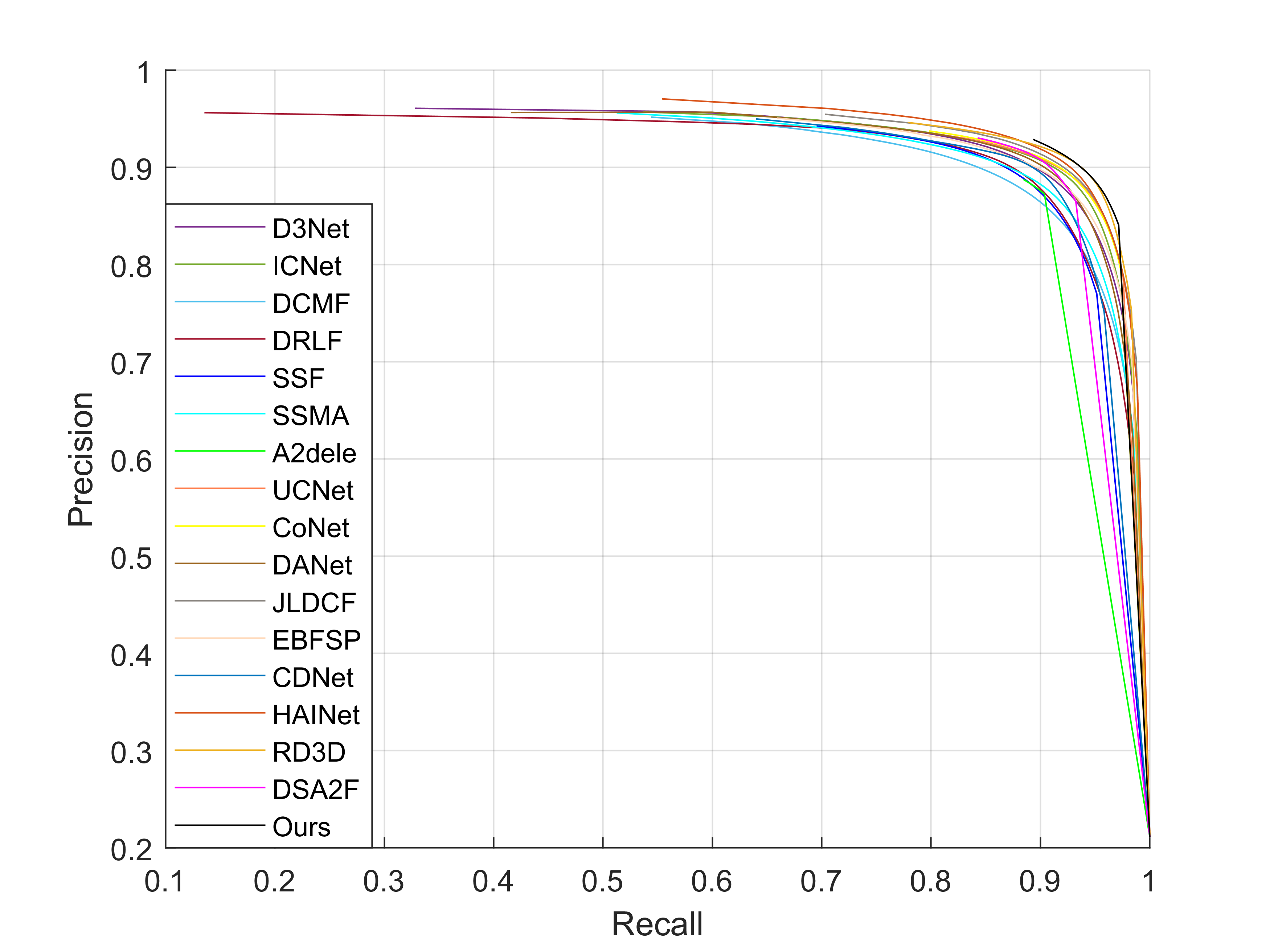}\\
(a)NLPR dataset&(b)NJU2K dataset&(c)STERE dataset\\
\end{tabular}
\begin{tabular}{ccc}
\includegraphics[width = 0.32\textwidth]{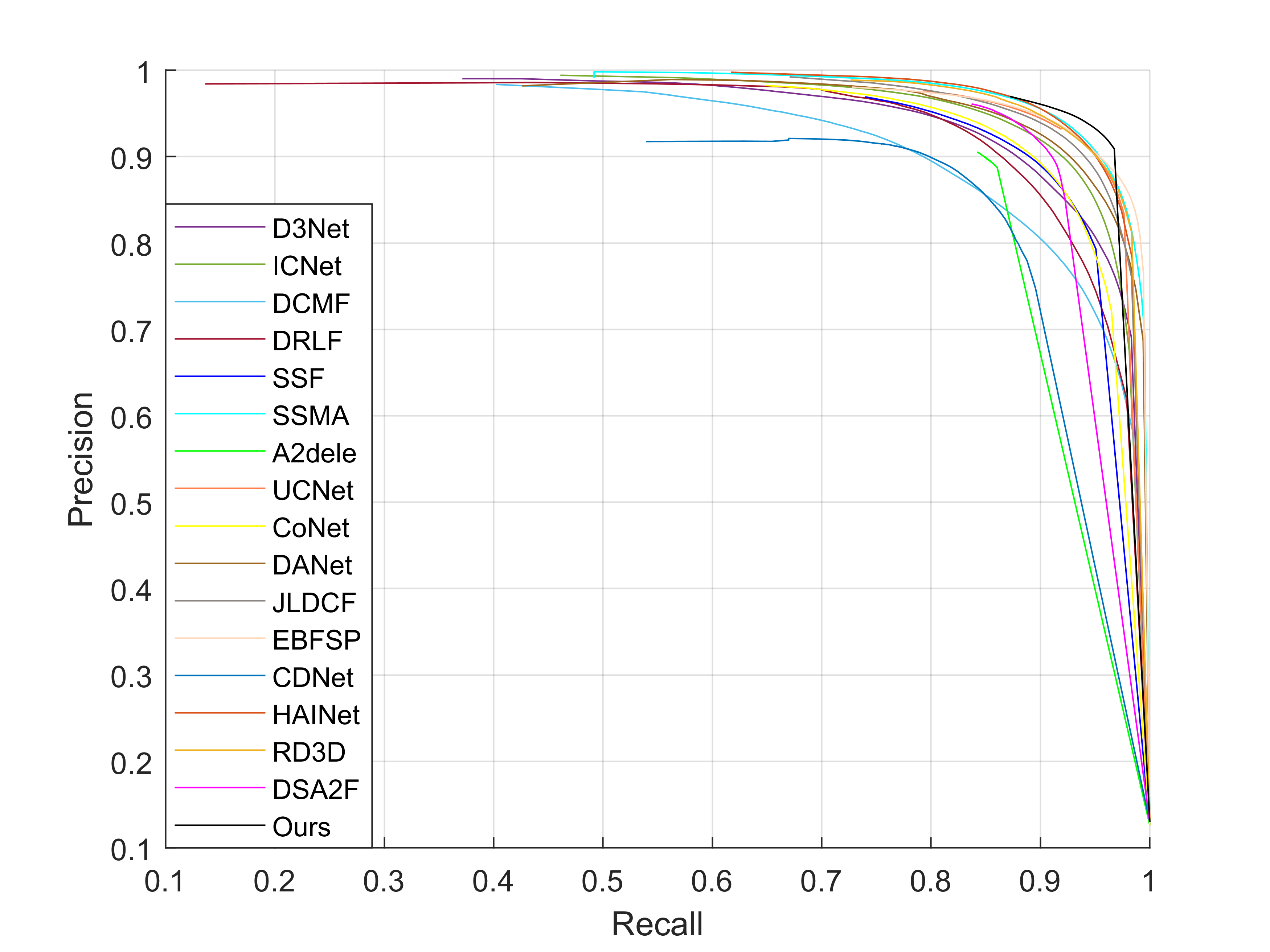}&\includegraphics[width = 0.32\textwidth]{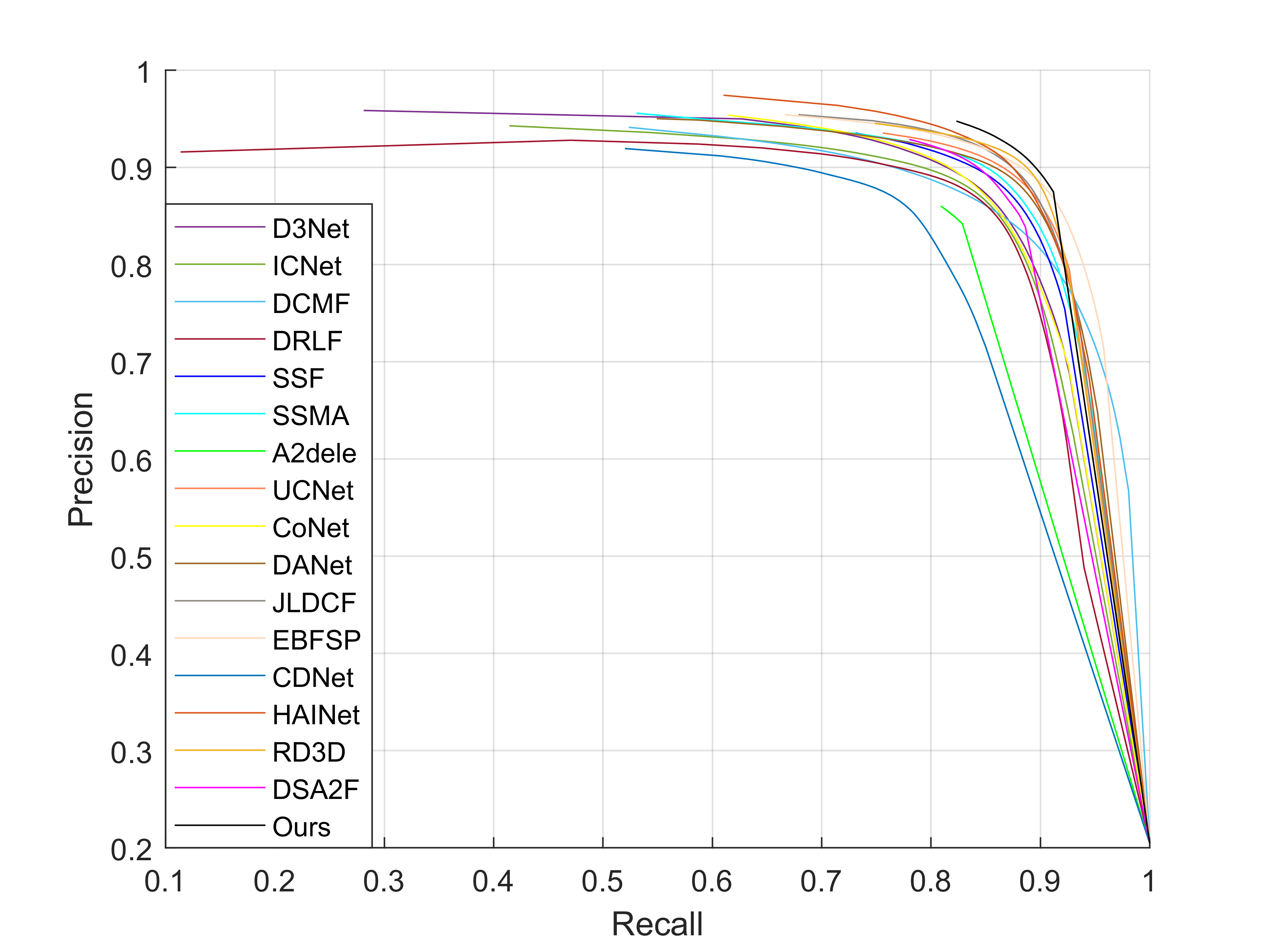}&\includegraphics[width = 0.32\textwidth]{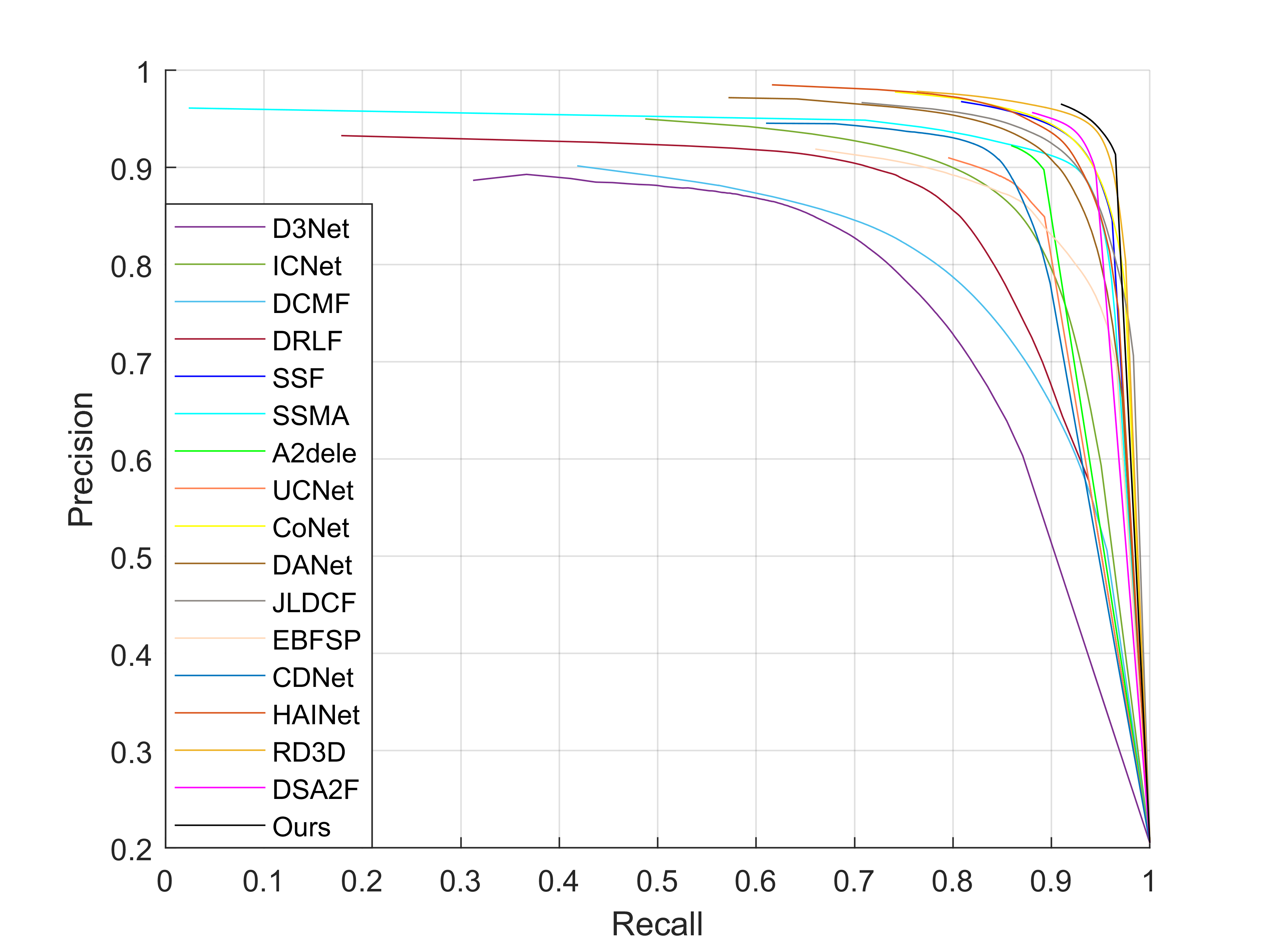}\\
(d)DES dataset&(e)SIP dataset&(f)DUT dataset\\
\end{tabular}
\caption{P-R curves comparisons of different models on six datasets.}
\label{fig:PRComparison}
\end{figure*}

Figure.\ref{fig:PRComparison} shows the comparison results on PR curve. Table.\ref{tab:comparison} shows the quantitative comparison results of four evaluation metrics.
As can be clearly observed from figure that our curves are very short, which means that our recall is very high. Furthermore, from the table, we can see that all the evaluation metrics are nearly the best on six datasets, so as to verify the effectiveness and advantages of our proposed method. Only two S-measure values in NLPR and STERE datasets are inferior to the best, but they are also the second best.
Combined with  the results of figure and table, our method achieves the impressive performance.

\begin{table*}[!htp]
  \centering
  \fontsize{6}{10}\selectfont
  \renewcommand{\arraystretch}{0.5}
  \renewcommand{\tabcolsep}{0.1mm}
  \scriptsize
  \caption{S-measure, adaptive F-measure, adaptive E-measure, MAE comparisons with different models.  The best result is in bold.}
\label{tab:comparison}
  \begin{tabular}{c|c|cccccccccccccccccc|c}
  \hline\toprule
    Datasets& Metric
          & D3Net   & ICNet & DCMF &DRLF  &SSF &SSMA &A2dele &UCNet &CoNet  &DANet &JLDCF &EBFSP&CDNet&HAINet& RD3D &DSA2F  &TriTransNet\\
       &  & TNNLS20   & TIP20 & TIP20 &TIP20  &CVPR20 &CVPR20 &CVPR20 &CVPR20 &ECCV20  &ECCV20 &CVPR20 &TMM21&TIP21& TIP21& AAAI21 &CVPR21 & Ours  \\

  \midrule

\multirow{3}{*}{\textit{NLPR}}
    & S$\uparrow$       & .912  & .923 & .900 &.903  &.914 &.915 &.896 &.920 &.908  &.920 &.925&.915&.902& .924 & \textbf{.930} &.918&.928\\

    & F$_\beta \uparrow$         & .861  & .870 & .839  &.843&.875&.853&.878 &.890 &.846 &.875 &.878&.897&.848& .897 & .892 &.892  &\textbf{.909} \\

    & $E_{\xi}\uparrow$       & .944& .944&.933&.936&.949&.938&.945 &.953 & .934  &.951 &.953&.952&.935& .957& .958&.950 &\textbf{.960}\\

    & MAE$\downarrow$           & .030  & .028 & .035 & .032 &.026&.030&.028 &.025 &.031 &.027 &.022&.026&.032& .024& .022&.024&\textbf{.020}\\
    \midrule

  \multirow{3}{*}{\textit{NJU2K}}
    & S$\uparrow$        & .901 & .894  & .889  & .886& .899     &.894&.869 &.897 &.895 &.899 &.902&.903&.885& .912 & .916 & .904 &\textbf{.920}\\

    &F$_\beta$$\uparrow$       & .865 & .868  & .859  & .849  & .886  &.865 &.874 &.889 &.872 &.871&.885&.894&.866 & .900 & .901 &.898 &\textbf{.919}\\

    & $E_{\xi}\uparrow$         & .914 & .905 & .897 & .901 & .913  & .896  & .897 &.903 & .912    &.908  &.913&.907&.911& .922 & .918&.922 &\textbf{.925}\\

    & MAE$\downarrow$               & .046  & .052 & .052 &.055&.043&.053& .051 &.043 & .046 &.045 &.041&.039&.048& .038 & .036&.039 &\textbf{.030}\\

    \midrule

  \multirow{3}{*}{\textit{STERE}}
    & S$\uparrow$       & .899 &.903  & .883   & .888 & .887 &.890 &.878 &.903 &.905 &.901&.903&.900&.896& .907 & \textbf{.911}& .897&.908\\

    & F$_\beta$$\uparrow$        & .859 & .865 & .841   & .845 & .867  &.855 &.874 &.885 &.884 &.868 &.869&.870&.873& .885 & .886 &\textbf{.893}&\textbf{.893}\\

    & $E_{\xi}\uparrow$         & .920 & .915 & .904 & .915 & .921  & .907 & .915 &.922 &\textbf{.927} &921 &.919&.912&.922& .925 & \textbf{.927}&\textbf{.927} &\textbf{.927}\\

    & MAE$\downarrow$             & .046 & .045  & .054    & .050 & .046 &.051 &.044 &.039 &.037 &.043 &.040&.045&.042& .040 & .037& .039&\textbf{.033}\\
    \midrule

  \multirow{3}{*}{\textit{DES}}
    & S$\uparrow$       & .898  & .920 & .877 &.895&.905 &.941&.885 &.933 &.911  &.924 &.931&.937&.875& .935 & .935&.916 &\textbf{.943}\\

    & F$_\beta$$\uparrow$        & .870 & .889  & .820   & .868 & .876&.906&.865 &.917 &.861&.899 &.900&.913&.839& .924 & .917&.901 &\textbf{.936}\\

    & $E_{\xi}\uparrow$          & .951 & .959 & .923 & .954 & .948 & .974   & .922 &.974 & .945  &.968 &.969&.974&.921& .974 & .975&.955&.\textbf{981}\\

    & MAE$\downarrow$              & .031 & .027  & .040   & .030 & .025  &.021&.028 &.018 &.027 &.023  &.020&.018&.034& .018 & .019&.023&\textbf{.014}\\
    \midrule

\multirow{3}{*}{\textit{SIP}}
    & S$\uparrow$      & .860  & .854 & .859 &.850 &.868 &.872 &.826 &.875 &.858 &.875&.880&.885&.823  & .880 & .885 &.862&\textbf{.886}\\

    & F$_\beta$$\uparrow$       & .835 & .836  & .819   & .813 & .851&.854&.825 &.868 &.842 &.855 &.873&.869&.805& .875 & .874& .865&\textbf{.892}\\

    & $E_{\xi}\uparrow$         & .902 & .899 & .898 & .891  & .911 & .911   & .892 &.913 & .909  &.914 &.921&.917&.880& .919 & .920& .908&\textbf{.924}\\

    & MAE$\downarrow$              & .063 & .069  & .068   & .071  & .056  &.057&.070 &.051 &.063 &.054 &.049&.049&.076& .053 & .048& .057&\textbf{.043}\\
    \midrule

  \multirow{3}{*}{\textit{DUT}}
    & S$\uparrow$       & .775  & .852 & .798 &.826&.916 &.903 &.886 &.864 &.919  &.899 &.906&.858&.880& .910 & .931 &.921&\textbf{.933}\\

    & F$_\beta$$\uparrow$       & .756 & .830  & .750   & .803 & .914&.866&.890 &.856 &.909 &.888  &.882&.842&.874& .906 & .924 &.926 &\textbf{.938}\\

    & $E_{\xi}\uparrow$         & .847 & .897 & .848 & .870  &946 & .921   & .924 &.903 &.948 &.934 &.931&.890&.918& .938 & .949 &.950&\textbf{.957}\\

    & MAE$\downarrow$               & .097 & .072  & .104   & .080 & .034  &.044&.043 &.056 &.033 &.043 &.043&.067&.048& .038 & .031 &.030 &\textbf{.025}\\

  \bottomrule
  \hline
  \end{tabular}

\end{table*}

\subsubsection{Qualitative Evaluation.} 
\begin{figure*}[!htp]
	\centering
	\includegraphics[width=1\textwidth]{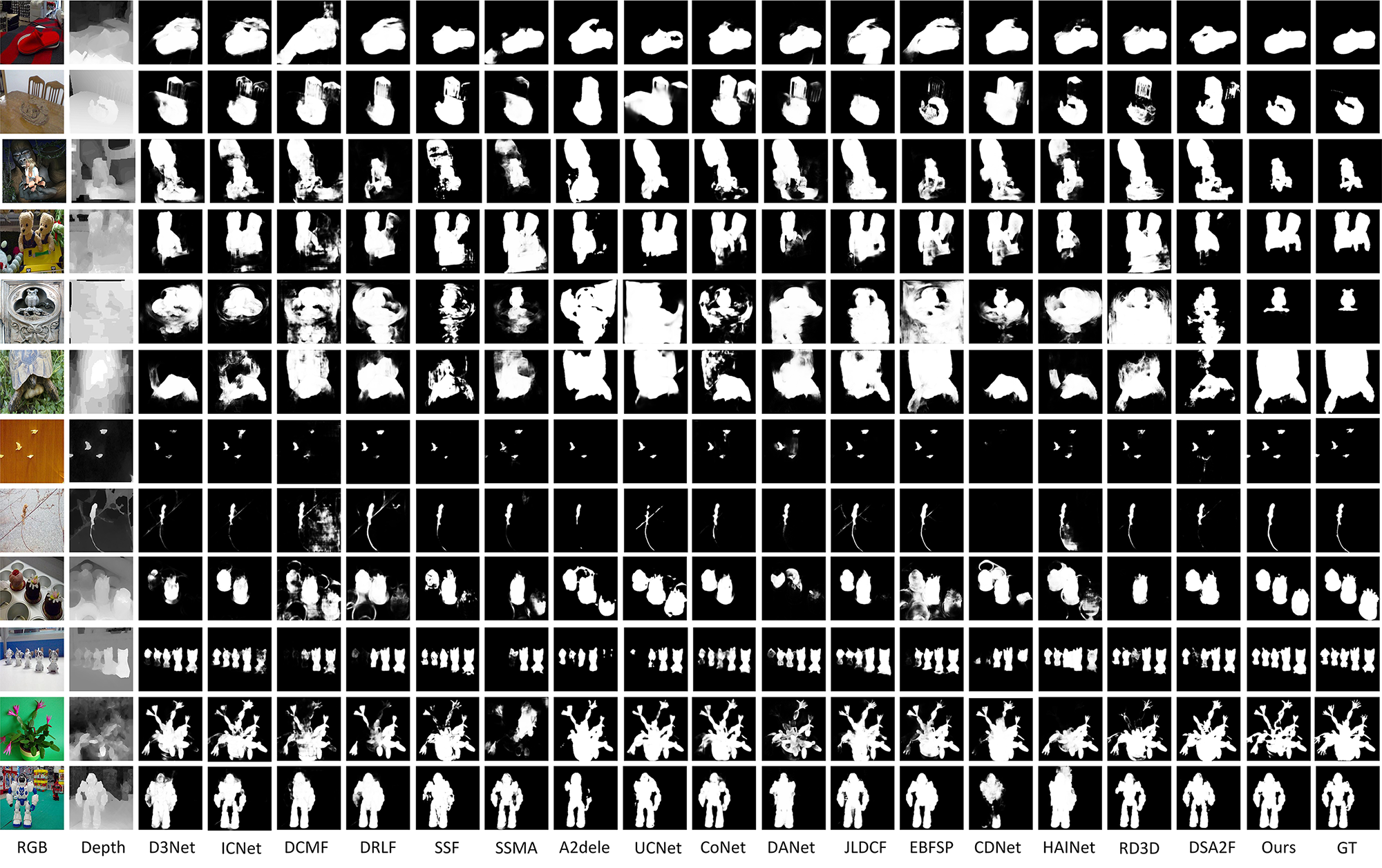}
	\caption{Visual comparison results with other the state-of-the-art models.  \label{visual_compare}}
\end{figure*}

To make the qualitative comparisons, we show some visual examples in Figure.\ref{visual_compare}.
It can be observed that our method has better detection results than other methods in some challenging cases: similar foreground and background($1^{st}_{}$-$2^{nd}_{}$ rows), complex scene($3^{rd}_{}$-$4^{th}_{}$ rows), low quality depth map($5^{th}_{}$-$6^{th}_{}$ rows), small object($7^{th}_{}$-$8^{th}_{}$ rows) and multiple objects($9^{th}_{}$-$10^{th}_{}$ rows). These indicate that our approach can better locate salient objects and produce more accurate saliency maps. In addition, our approach can produce more fine-grained details as highlighted in the salient region($11^{th}_{}$-$12^{th}_{}$ rows). This is also the proof of the effectiveness of our method.

\subsection{Ablation studies}
We conduct ablation studies on NLPR, NJU2K, SIP and STERE  datasets to investigate the contributions of different modules in the proposed method.
\subsubsection{The effectiveness of triplet transformer embedding module (TTEM)}
The baseline model used here removes TTEM. Its performance is shown in the variant No.1 of Table.~\ref{tab:Ablation1}.
Further, we replace TTEM with gated recurrent unit (GRU)~\cite{cho2014learning}, whose result is shown in the variant No.2 of Table.~\ref{tab:Ablation1}.
The variant No.3 of Table.~\ref{tab:Ablation1} is the result of siamese transformer applied in the high two layers.
The variant No.4 of Table.~\ref{tab:Ablation1} is the result of quadruplet transformer applied in the high four layers.
The variant No.5 of Table.~\ref{tab:Ablation1} is our result of triplet transformer applied in the high three layers.

It can be clearly observed that compared with No.1,  the result of our TriTransNet is improved $0.016$  in the S-measure metric, $0.021$  in the F-measure metric, $0.008$  in the E-measure metric and $0.007$  in the MAE metric on average.
Meanwhile, compared with No.2, the result of our TriTransNet is improved  $0.012$ in the S-measure metric, $0.014$ in the F-measure metric, $0.006$ in the E-measure metric and  $0.005$ in the MAE metric on average. TTEM plays an important role in the performance improvement.

In addition, we compare No.3, No.4 and No.5 and find that Triplet win Siamese in S-measure, F-measure, E-measure, and MAE about 0.009,0.016,0.006 and 0.005 on average, and outperform Quadruplet about 0.010,0.009,0.005 and 0.004 on average. Our TriTransNet enhances long-range dependency of semantic information by using the features in the high three layers, and further combines with three-stream usampling decoding in the low two layers to perfectly depict the detailed boundary, so as to achieve the best performance.
\begin{table*}[!htp]
    \centering
    \fontsize{6}{10}\selectfont
    \setlength{\tabcolsep}{0.3mm}{
    \caption{Ablation experiment of triplet transformer embedding module (TTEM). The best result is in bold.}
    \label{tab:Ablation1}
\begin{tabular}{c|ccccc|cccc|cccc|cccc|cccc}
    \hline\toprule
   \multirow{2}{*}{\centering Variant} & \multicolumn{5}{c|}{\centering Candidate} & \multicolumn{4}{c|}{\centering NLPR} & \multicolumn{4}{c|}{\centering NJUD2K} & \multicolumn{4}{c|}{\centering SIP} & \multicolumn{4}{c}{\centering STERE}\\
     &  Baseline &GRU &Siamese&Quadruplet&Triplet
     &S$\uparrow$& F$_\beta$ $\uparrow$ &$E_{\xi}\uparrow$
     & MAE$\downarrow$
     &S$\uparrow$& F$_\beta$ $\uparrow$ &$E_{\xi}\uparrow$
     & MAE$\downarrow$
      &S$\uparrow$& F$_\beta$$\uparrow$&$E_{\xi}\uparrow$ & MAE$\downarrow$
      &S$\uparrow$& F$_\beta$$\uparrow$ &$E_{\xi}\uparrow$& MAE$\downarrow$ \\
    \hline

    No.1  &$\checkmark$ & &&&
     &.910&.882& .952 &.026
      &.904&.897& .917 &.038
       &.876&.877& .918 &.049
        &.888&.873& .916 &.042 \\
    No.2  &$\checkmark$ &$\checkmark$ &&&
    &.914&.891& .953 &.024
     &.905&.901& .919 &.037
      &.879&.882& .919 &.047
      &.895&.883& .920&.038 \\
        No.3  &$\checkmark$ & &$\checkmark$&&
    &.917&.888& .956 &.024
     &.910&.903& .915 &.035
      &.882&.885& .926 &.046
      &.896&.872& .917&.040 \\
        No.4  &$\checkmark$ & & &$\checkmark$&
    &.922&.903& .958 &.022
     &.911&.908& .922 &.034
      &.875&.886& .913 &.048
      &.895&.881& .922&.038 \\
    No.5  &$\checkmark$ & &&&$\checkmark$
      &\textbf{.928}&\textbf{.909}&\textbf{ .960} &\textbf{.020}
      &\textbf{.920}&\textbf{.919}&\textbf{ .925}&\textbf{.030}
      &\textbf{.886}&\textbf{.892}&\textbf{ .924} &\textbf{.043}
       &\textbf{.908}&\textbf{.893}&\textbf{ .927} &\textbf{.033} \\
    \bottomrule
    \hline
\end{tabular}}
\end{table*}
\subsubsection{The effectiveness of three-stream decoder}
we further conduct the ablation study by replacing three stream decoder with single-stream decoder to check the effectiveness of the designed three-stream decoder. Table.~\ref{tab:Ablation2} No.1 denotes the model which adopts single-stream decoder and No.2 means our three-stream decoder. From Table.~\ref{tab:Ablation2}, we can see that the use of three-stream decoder obviously improves the detection performance. It benefits from the full integration of multi-layer features.
\begin{table*}[!htp]
    \centering
    \fontsize{6}{10}\selectfont
    \setlength{\tabcolsep}{0.3mm}{
    \caption{Ablation experiment of three-stream decoder. The best result is in bold.}
    \label{tab:Ablation2}
\begin{tabular}{c|cc|cccc|cccc|cccc|cccc}
    \hline\toprule
   \multirow{2}{*}{\centering Variant} & \multicolumn{2}{c|}{\centering Candidate} & \multicolumn{4}{c|}{\centering NLPR} & \multicolumn{4}{c|}{\centering NJUD2K} & \multicolumn{4}{c|}{\centering SIP} & \multicolumn{4}{c}{\centering STERE}\\
     & single-stream &three-stream  
     &S$\uparrow$& F$_\beta$ $\uparrow$ &$E_{\xi}\uparrow$
     & MAE$\downarrow$ 
     &S$\uparrow$& F$_\beta$ $\uparrow$ &$E_{\xi}\uparrow$
     & MAE$\downarrow$ 
      &S$\uparrow$& F$_\beta$$\uparrow$&$E_{\xi}\uparrow$ & MAE$\downarrow$ 
      &S$\uparrow$& F$_\beta$$\uparrow$ &$E_{\xi}\uparrow$& MAE$\downarrow$ \\
    \hline

    No.1  &$\checkmark$ &   
     &.923&.892& .958 &.022
      &.916&.904& .919 &.035 
       &.884&.886& .920 &.045 
        &.903&.879& .920 &.037 \\
    No.2  & &$\checkmark$ 
     &\textbf{.928}&\textbf{.909}&\textbf{ .960} &\textbf{.020}
      &\textbf{.920}&\textbf{.919}&\textbf{ .925}&\textbf{.030} 
      &\textbf{.886}&\textbf{.892}&\textbf{ .924} &\textbf{.043} 
       &\textbf{.908}&\textbf{.893}&\textbf{ .927} &\textbf{.033} \\
    \bottomrule
    \hline
\end{tabular}}
\end{table*}
\subsubsection{The effectiveness of depth purification module (DPM)}
The baseline model used here removes depth purification module (DPM). It attaches the depth feature to color feature by element-wise addition operation in the encoder. Its performance is illustrated in the variant No.1 of Table.~\ref{tab:Ablation3}.
Further, we discuss the similar depth-enhanced module (DEM) proposed in BBS\cite{fan2020bbs} whose result is shown in the variant No.2 of Table.~\ref{tab:Ablation3}. The variant No.3 of Table.~\ref{tab:Ablation3} denotes the model which adopts DPM instead of element-wise addition operation based on the baseline.

Compared with No.1, the performance of the variant No.3 is significantly improved. Meanwhile,
compared with No.2 which using DEM, our detection effect is also  better than that of No.2.
It verified that the effectiveness of DPM.
\begin{table*}[!htp]
    \centering
    \fontsize{6}{10}\selectfont
    \setlength{\tabcolsep}{0.5mm}{
    \caption{Ablation experiment of depth purification module (DPM). The best result is in bold.}
    \label{tab:Ablation3}
\begin{tabular}{c|ccc|cccc|cccc|cccc|cccc}
    \hline\toprule
   \multirow{2}{*}{\centering Variant} & \multicolumn{3}{c|}{\centering Candidate} & \multicolumn{4}{c|}{\centering NLPR} & \multicolumn{4}{c|}{\centering NJUD2K} & \multicolumn{4}{c|}{\centering SIP} & \multicolumn{4}{c}{\centering STERE}\\
     &  Baseline &DEM &DPM 
     &S$\uparrow$& F$_\beta$$\uparrow$ &$E_{\xi}\uparrow$
     & MAE$\downarrow$ 
     &S$\uparrow$& F$_\beta$$\uparrow$ &$E_{\xi}\uparrow$
     & MAE$\downarrow$ 
      &S$\uparrow$& F$_\beta$$\uparrow$&$E_{\xi}\uparrow$ & MAE$\downarrow$ 
      &S$\uparrow$& F$_\beta$$\uparrow$ &$E_{\xi}\uparrow$& MAE$\downarrow$ \\
    \hline

    No.1  &$\checkmark$ & &  
     &.917&.897& .956 &.023
      &.909&.904& .920 &.035 
       &.883&.887& .921 &.044 
        &.894&.875& .918 &.039 \\
    No.2  &$\checkmark$ &$\checkmark$ &  
    &.923&.901& .958 &.021
     &.914&.910& .922 &.033
      &.884&.889& .923 &.044 
      &.905&.889& .925&.035 \\
    No.3  &$\checkmark$ & &$\checkmark$ 
     &\textbf{.928}&\textbf{.909}&\textbf{ .960} &\textbf{.020}
      &\textbf{.920}&\textbf{.919}&\textbf{ .925}&\textbf{.030} 
      &\textbf{.886}&\textbf{.892}&\textbf{ .924} &\textbf{.043} 
       &\textbf{.908}&\textbf{.893}&\textbf{ .927} &\textbf{.033} \\
    \bottomrule
    \hline
\end{tabular}}
\end{table*}

\section{Conclusions}
In the paper, we introduce transformer into U-Net framework to detect salient object in RGB-D image.
Different from existing combination method of transformer and convolutional neural networks, we propose a triplet transformer embedding module which can be embedded into existing U-Net models for the better feature representation by learning long-range dependency among different levels with less cost.
Furthermore, we use depth information to enhance RGB features by depth purification module.
Experimental results show our method pushes the performance to a new level, and ablation studies also verify the effectiveness of each module. In the future, we will achieve the same task by a pure transformer, and further discuss their respective advantages to achieve the better combination.

\begin{acks}
This work is supported by National Natural Science Foundation of China (62006002),  Natural Science Foundation of Anhui Province (1908085MF182) and Key Program of Natural Science Project of Educational Commission of Anhui Province(KJ2019A0034).
\end{acks}
\bibliographystyle{ACM-Reference-Format}
\balance
\bibliography{TriTransNet}


\begin{thebibliography}{94}


\ifx \showCODEN    \undefined \def \showCODEN     #1{\unskip}     \fi
\ifx \showDOI      \undefined \def \showDOI       #1{#1}\fi
\ifx \showISBNx    \undefined \def \showISBNx     #1{\unskip}     \fi
\ifx \showISBNxiii \undefined \def \showISBNxiii  #1{\unskip}     \fi
\ifx \showISSN     \undefined \def \showISSN      #1{\unskip}     \fi
\ifx \showLCCN     \undefined \def \showLCCN      #1{\unskip}     \fi
\ifx \shownote     \undefined \def \shownote      #1{#1}          \fi
\ifx \showarticletitle \undefined \def \showarticletitle #1{#1}   \fi
\ifx \showURL      \undefined \def \showURL       {\relax}        \fi
\providecommand\bibfield[2]{#2}
\providecommand\bibinfo[2]{#2}
\providecommand\natexlab[1]{#1}
\providecommand\showeprint[2][]{arXiv:#2}

\bibitem[\protect\citeauthoryear{Achanta, Hemami, Estrada, and
  Susstrunk}{Achanta et~al\mbox{.}}{2009}]%
        {achanta2009frequency}
\bibfield{author}{\bibinfo{person}{Radhakrishna Achanta},
  \bibinfo{person}{Sheila Hemami}, \bibinfo{person}{Francisco Estrada}, {and}
  \bibinfo{person}{Sabine Susstrunk}.} \bibinfo{year}{2009}\natexlab{}.
\newblock \showarticletitle{{Frequency-tuned salient region detection}}. In
  \bibinfo{booktitle}{\emph{2009 IEEE conference on computer vision and pattern
  recognition}}. IEEE, \bibinfo{pages}{1597--1604}.
\newblock


\bibitem[\protect\citeauthoryear{Ba, Kiros, and Hinton}{Ba
  et~al\mbox{.}}{2016}]%
        {ba2016layer}
\bibfield{author}{\bibinfo{person}{Jimmy~Lei Ba}, \bibinfo{person}{Jamie~Ryan
  Kiros}, {and} \bibinfo{person}{Geoffrey~E Hinton}.}
  \bibinfo{year}{2016}\natexlab{}.
\newblock \showarticletitle{Layer normalization}.
\newblock \bibinfo{journal}{\emph{arXiv preprint arXiv:1607.06450}}
  (\bibinfo{year}{2016}).
\newblock


\bibitem[\protect\citeauthoryear{Borji, Cheng, Jiang, and Li}{Borji
  et~al\mbox{.}}{2015}]%
        {borji2015salient}
\bibfield{author}{\bibinfo{person}{Ali Borji}, \bibinfo{person}{Ming-Ming
  Cheng}, \bibinfo{person}{Huaizu Jiang}, {and} \bibinfo{person}{Jia Li}.}
  \bibinfo{year}{2015}\natexlab{}.
\newblock \showarticletitle{{Salient object detection: A benchmark}}.
\newblock \bibinfo{journal}{\emph{IEEE transactions on image processing}}
  \bibinfo{volume}{24}, \bibinfo{number}{12} (\bibinfo{year}{2015}),
  \bibinfo{pages}{5706--5722}.
\newblock


\bibitem[\protect\citeauthoryear{Carion, Massa, Synnaeve, Usunier, Kirillov,
  and Zagoruyko}{Carion et~al\mbox{.}}{2020}]%
        {carion2020end}
\bibfield{author}{\bibinfo{person}{Nicolas Carion}, \bibinfo{person}{Francisco
  Massa}, \bibinfo{person}{Gabriel Synnaeve}, \bibinfo{person}{Nicolas
  Usunier}, \bibinfo{person}{Alexander Kirillov}, {and} \bibinfo{person}{Sergey
  Zagoruyko}.} \bibinfo{year}{2020}\natexlab{}.
\newblock \showarticletitle{{End-to-end object detection with transformers}}.
  In \bibinfo{booktitle}{\emph{European Conference on Computer Vision}}.
  Springer, \bibinfo{pages}{213--229}.
\newblock


\bibitem[\protect\citeauthoryear{Chen, Wei, Peng, and Qin}{Chen
  et~al\mbox{.}}{2021c}]%
        {chen2020depth}
\bibfield{author}{\bibinfo{person}{Chenglizhao Chen}, \bibinfo{person}{Jipeng
  Wei}, \bibinfo{person}{Chong Peng}, {and} \bibinfo{person}{Hong Qin}.}
  \bibinfo{year}{2021}\natexlab{c}.
\newblock \showarticletitle{Depth-Quality-Aware Salient Object Detection}.
\newblock \bibinfo{journal}{\emph{IEEE Transactions on Image Processing}}
  \bibinfo{volume}{30} (\bibinfo{year}{2021}), \bibinfo{pages}{2350--2363}.
\newblock


\bibitem[\protect\citeauthoryear{Chen, Deng, Li, Hung, and Lin}{Chen
  et~al\mbox{.}}{2020b}]%
        {chen2020rgbd}
\bibfield{author}{\bibinfo{person}{Hao Chen}, \bibinfo{person}{Yongjian Deng},
  \bibinfo{person}{Youfu Li}, \bibinfo{person}{Tzu-Yi Hung}, {and}
  \bibinfo{person}{Guosheng Lin}.} \bibinfo{year}{2020}\natexlab{b}.
\newblock \showarticletitle{{RGBD salient object detection via disentangled
  cross-modal fusion}}.
\newblock \bibinfo{journal}{\emph{IEEE Transactions on Image Processing}}
  \bibinfo{volume}{29} (\bibinfo{year}{2020}), \bibinfo{pages}{8407--8416}.
\newblock


\bibitem[\protect\citeauthoryear{Chen and Li}{Chen and Li}{2018}]%
        {chen2018progressively}
\bibfield{author}{\bibinfo{person}{Hao Chen} {and} \bibinfo{person}{Youfu Li}.}
  \bibinfo{year}{2018}\natexlab{}.
\newblock \showarticletitle{{Progressively complementarity-aware fusion network
  for RGB-D salient object detection}}. In
  \bibinfo{booktitle}{\emph{Proceedings of the IEEE conference on computer
  vision and pattern recognition}}. \bibinfo{pages}{3051--3060}.
\newblock


\bibitem[\protect\citeauthoryear{Chen, Lu, Yu, Luo, Adeli, Wang, Lu, Yuille,
  and Zhou}{Chen et~al\mbox{.}}{2021b}]%
        {chen2021transunet}
\bibfield{author}{\bibinfo{person}{Jieneng Chen}, \bibinfo{person}{Yongyi Lu},
  \bibinfo{person}{Qihang Yu}, \bibinfo{person}{Xiangde Luo},
  \bibinfo{person}{Ehsan Adeli}, \bibinfo{person}{Yan Wang},
  \bibinfo{person}{Le Lu}, \bibinfo{person}{Alan~L Yuille}, {and}
  \bibinfo{person}{Yuyin Zhou}.} \bibinfo{year}{2021}\natexlab{b}.
\newblock \showarticletitle{{Transunet: Transformers make strong encoders for
  medical image segmentation}}.
\newblock \bibinfo{journal}{\emph{arXiv preprint arXiv:2102.04306}}
  (\bibinfo{year}{2021}).
\newblock


\bibitem[\protect\citeauthoryear{Chen, Fu, Liu, Chen, Du, Qiu, and Shao}{Chen
  et~al\mbox{.}}{2020c}]%
        {chen2020ef}
\bibfield{author}{\bibinfo{person}{Qian Chen}, \bibinfo{person}{Keren Fu},
  \bibinfo{person}{Ze Liu}, \bibinfo{person}{Geng Chen},
  \bibinfo{person}{Hongwei Du}, \bibinfo{person}{Bensheng Qiu}, {and}
  \bibinfo{person}{Ling Shao}.} \bibinfo{year}{2020}\natexlab{c}.
\newblock \showarticletitle{{EF-Net: A novel enhancement and fusion network for
  RGB-D saliency detection}}.
\newblock \bibinfo{journal}{\emph{Pattern Recognition}} (\bibinfo{year}{2020}),
  \bibinfo{pages}{107740}.
\newblock


\bibitem[\protect\citeauthoryear{Chen, Liu, Zhang, Fu, Zhao, and Du}{Chen
  et~al\mbox{.}}{2021a}]%
        {chen2021rd3d}
\bibfield{author}{\bibinfo{person}{Qian Chen}, \bibinfo{person}{Ze Liu},
  \bibinfo{person}{Yi Zhang}, \bibinfo{person}{Keren Fu},
  \bibinfo{person}{Qijun Zhao}, {and} \bibinfo{person}{Hongwei Du}.}
  \bibinfo{year}{2021}\natexlab{a}.
\newblock \showarticletitle{RGB-D Salient Object Detection via 3D Convolutional
  Neural}.
\newblock \bibinfo{journal}{\emph{AAAI}} (\bibinfo{year}{2021}).
\newblock


\bibitem[\protect\citeauthoryear{Chen and Fu}{Chen and Fu}{2020}]%
        {chen2020progressively}
\bibfield{author}{\bibinfo{person}{Shuhan Chen} {and} \bibinfo{person}{Yun
  Fu}.} \bibinfo{year}{2020}\natexlab{}.
\newblock \showarticletitle{{Progressively guided alternate refinement network
  for RGB-D salient object detection}}. In \bibinfo{booktitle}{\emph{European
  Conference on Computer Vision}}. Springer, \bibinfo{pages}{520--538}.
\newblock


\bibitem[\protect\citeauthoryear{Chen, Zhu, Liu, He, and Liu}{Chen
  et~al\mbox{.}}{2021e}]%
        {chen2021global}
\bibfield{author}{\bibinfo{person}{Sihan Chen}, \bibinfo{person}{Xinxin Zhu},
  \bibinfo{person}{Wei Liu}, \bibinfo{person}{Xingjian He}, {and}
  \bibinfo{person}{Jing Liu}.} \bibinfo{year}{2021}\natexlab{e}.
\newblock \showarticletitle{{Global-Local Propagation Network for RGB-D
  Semantic Segmentation}}.
\newblock \bibinfo{journal}{\emph{arXiv preprint arXiv:2101.10801}}
  (\bibinfo{year}{2021}).
\newblock


\bibitem[\protect\citeauthoryear{Chen, Yan, Zhu, Wang, Yang, and Lu}{Chen
  et~al\mbox{.}}{2021d}]%
        {chen2021transformer}
\bibfield{author}{\bibinfo{person}{Xin Chen}, \bibinfo{person}{Bin Yan},
  \bibinfo{person}{Jiawen Zhu}, \bibinfo{person}{Dong Wang},
  \bibinfo{person}{Xiaoyun Yang}, {and} \bibinfo{person}{Huchuan Lu}.}
  \bibinfo{year}{2021}\natexlab{d}.
\newblock \showarticletitle{Transformer tracking}. In
  \bibinfo{booktitle}{\emph{Proceedings of the IEEE/CVF Conference on Computer
  Vision and Pattern Recognition}}. \bibinfo{pages}{8126--8135}.
\newblock


\bibitem[\protect\citeauthoryear{Chen, Cong, Xu, and Huang}{Chen
  et~al\mbox{.}}{2020a}]%
        {chen2020dpanet}
\bibfield{author}{\bibinfo{person}{Zuyao Chen}, \bibinfo{person}{Runmin Cong},
  \bibinfo{person}{Qianqian Xu}, {and} \bibinfo{person}{Qingming Huang}.}
  \bibinfo{year}{2020}\natexlab{a}.
\newblock \showarticletitle{{DPANet: Depth Potentiality-Aware Gated Attention
  Network for RGB-D Salient Object Detection}}.
\newblock \bibinfo{journal}{\emph{IEEE Transactions on Image Processing}}
  (\bibinfo{year}{2020}).
\newblock


\bibitem[\protect\citeauthoryear{Cheng, Fu, Wei, Xiao, and Cao}{Cheng
  et~al\mbox{.}}{2014}]%
        {cheng2014depth}
\bibfield{author}{\bibinfo{person}{Yupeng Cheng}, \bibinfo{person}{Huazhu Fu},
  \bibinfo{person}{Xingxing Wei}, \bibinfo{person}{Jiangjian Xiao}, {and}
  \bibinfo{person}{Xiaochun Cao}.} \bibinfo{year}{2014}\natexlab{}.
\newblock \showarticletitle{{Depth enhanced saliency detection method}}. In
  \bibinfo{booktitle}{\emph{Proceedings of international conference on internet
  multimedia computing and service}}. \bibinfo{pages}{23--27}.
\newblock


\bibitem[\protect\citeauthoryear{Cho, Van~Merri{\"e}nboer, Gulcehre, Bahdanau,
  Bougares, Schwenk, and Bengio}{Cho et~al\mbox{.}}{2014}]%
        {cho2014learning}
\bibfield{author}{\bibinfo{person}{Kyunghyun Cho}, \bibinfo{person}{Bart
  Van~Merri{\"e}nboer}, \bibinfo{person}{Caglar Gulcehre},
  \bibinfo{person}{Dzmitry Bahdanau}, \bibinfo{person}{Fethi Bougares},
  \bibinfo{person}{Holger Schwenk}, {and} \bibinfo{person}{Yoshua Bengio}.}
  \bibinfo{year}{2014}\natexlab{}.
\newblock \showarticletitle{{Learning phrase representations using RNN
  encoder-decoder for statistical machine translation}}.
\newblock \bibinfo{journal}{\emph{arXiv preprint arXiv:1406.1078}}
  (\bibinfo{year}{2014}).
\newblock


\bibitem[\protect\citeauthoryear{Chu, Zhang, Tian, Wei, and Xia}{Chu
  et~al\mbox{.}}{2021}]%
        {chu2021we}
\bibfield{author}{\bibinfo{person}{Xiangxiang Chu}, \bibinfo{person}{Bo Zhang},
  \bibinfo{person}{Zhi Tian}, \bibinfo{person}{Xiaolin Wei}, {and}
  \bibinfo{person}{Huaxia Xia}.} \bibinfo{year}{2021}\natexlab{}.
\newblock \showarticletitle{{Do We Really Need Explicit Position Encodings for
  Vision Transformers?}}
\newblock \bibinfo{journal}{\emph{arXiv preprint arXiv:2102.10882}}
  (\bibinfo{year}{2021}).
\newblock


\bibitem[\protect\citeauthoryear{Donoser, Urschler, Hirzer, and
  Bischof}{Donoser et~al\mbox{.}}{2009}]%
        {donoser2009saliency}
\bibfield{author}{\bibinfo{person}{Michael Donoser}, \bibinfo{person}{Martin
  Urschler}, \bibinfo{person}{Martin Hirzer}, {and} \bibinfo{person}{Horst
  Bischof}.} \bibinfo{year}{2009}\natexlab{}.
\newblock \showarticletitle{{Saliency driven total variation segmentation}}. In
  \bibinfo{booktitle}{\emph{2009 IEEE 12th International Conference on Computer
  Vision}}. IEEE, \bibinfo{pages}{817--824}.
\newblock


\bibitem[\protect\citeauthoryear{Dosovitskiy, Beyer, Kolesnikov, Weissenborn,
  Zhai, Unterthiner, Dehghani, Minderer, Heigold, Gelly, Uszkoreit, and
  Houlsby}{Dosovitskiy et~al\mbox{.}}{2021}]%
        {dosovitskiy2021an}
\bibfield{author}{\bibinfo{person}{Alexey Dosovitskiy}, \bibinfo{person}{Lucas
  Beyer}, \bibinfo{person}{Alexander Kolesnikov}, \bibinfo{person}{Dirk
  Weissenborn}, \bibinfo{person}{Xiaohua Zhai}, \bibinfo{person}{Thomas
  Unterthiner}, \bibinfo{person}{Mostafa Dehghani}, \bibinfo{person}{Matthias
  Minderer}, \bibinfo{person}{Georg Heigold}, \bibinfo{person}{Sylvain Gelly},
  \bibinfo{person}{Jakob Uszkoreit}, {and} \bibinfo{person}{Neil Houlsby}.}
  \bibinfo{year}{2021}\natexlab{}.
\newblock \showarticletitle{An Image is Worth 16x16 Words: Transformers for
  Image Recognition at Scale}. In \bibinfo{booktitle}{\emph{International
  Conference on Learning Representations}}.
\newblock


\bibitem[\protect\citeauthoryear{Fan, Cheng, Liu, Li, and Borji}{Fan
  et~al\mbox{.}}{2017}]%
        {fan2017structure}
\bibfield{author}{\bibinfo{person}{Deng-Ping Fan}, \bibinfo{person}{Ming-Ming
  Cheng}, \bibinfo{person}{Yun Liu}, \bibinfo{person}{Tao Li}, {and}
  \bibinfo{person}{Ali Borji}.} \bibinfo{year}{2017}\natexlab{}.
\newblock \showarticletitle{{Structure-measure: A new way to evaluate
  foreground maps}}. In \bibinfo{booktitle}{\emph{Proceedings of the IEEE
  international conference on computer vision}}. \bibinfo{pages}{4548--4557}.
\newblock


\bibitem[\protect\citeauthoryear{Fan, Gong, Cao, Ren, Cheng, and Borji}{Fan
  et~al\mbox{.}}{2018}]%
        {fan2018enhanced}
\bibfield{author}{\bibinfo{person}{Deng-Ping Fan}, \bibinfo{person}{Cheng
  Gong}, \bibinfo{person}{Yang Cao}, \bibinfo{person}{Bo Ren},
  \bibinfo{person}{Ming-Ming Cheng}, {and} \bibinfo{person}{Ali Borji}.}
  \bibinfo{year}{2018}\natexlab{}.
\newblock \showarticletitle{{Enhanced-alignment measure for binary foreground
  map evaluation}}.
\newblock \bibinfo{journal}{\emph{arXiv preprint arXiv:1805.10421}}
  (\bibinfo{year}{2018}).
\newblock


\bibitem[\protect\citeauthoryear{Fan, Lin, Zhang, Zhu, and Cheng}{Fan
  et~al\mbox{.}}{2020a}]%
        {fan2020rethinking}
\bibfield{author}{\bibinfo{person}{Deng-Ping Fan}, \bibinfo{person}{Zheng Lin},
  \bibinfo{person}{Zhao Zhang}, \bibinfo{person}{Menglong Zhu}, {and}
  \bibinfo{person}{Ming-Ming Cheng}.} \bibinfo{year}{2020}\natexlab{a}.
\newblock \showarticletitle{{Rethinking RGB-D Salient Object Detection: Models,
  Data Sets, and Large-Scale Benchmarks}}.
\newblock \bibinfo{journal}{\emph{IEEE Transactions on Neural Networks and
  Learning Systems}} (\bibinfo{year}{2020}).
\newblock


\bibitem[\protect\citeauthoryear{Fan, Zhai, Borji, Yang, and Shao}{Fan
  et~al\mbox{.}}{2020b}]%
        {fan2020bbs}
\bibfield{author}{\bibinfo{person}{Deng-Ping Fan}, \bibinfo{person}{Yingjie
  Zhai}, \bibinfo{person}{Ali Borji}, \bibinfo{person}{Jufeng Yang}, {and}
  \bibinfo{person}{Ling Shao}.} \bibinfo{year}{2020}\natexlab{b}.
\newblock \showarticletitle{{BBS-Net: RGB-D salient object detection with a
  bifurcated backbone strategy network}}. In \bibinfo{booktitle}{\emph{European
  Conference on Computer Vision}}. Springer, \bibinfo{pages}{275--292}.
\newblock


\bibitem[\protect\citeauthoryear{Fu, Fan, Ji, and Zhao}{Fu
  et~al\mbox{.}}{2020}]%
        {fu2020jl}
\bibfield{author}{\bibinfo{person}{Keren Fu}, \bibinfo{person}{Deng-Ping Fan},
  \bibinfo{person}{Ge-Peng Ji}, {and} \bibinfo{person}{Qijun Zhao}.}
  \bibinfo{year}{2020}\natexlab{}.
\newblock \showarticletitle{{JL-DCF: Joint learning and densely-cooperative
  fusion framework for rgb-d salient object detection}}. In
  \bibinfo{booktitle}{\emph{Proceedings of the IEEE/CVF conference on computer
  vision and pattern recognition}}. \bibinfo{pages}{3052--3062}.
\newblock


\bibitem[\protect\citeauthoryear{Gao, Shi, Tao, and Xu}{Gao
  et~al\mbox{.}}{2015}]%
        {gao2015database}
\bibfield{author}{\bibinfo{person}{Yuan Gao}, \bibinfo{person}{Miaojing Shi},
  \bibinfo{person}{Dacheng Tao}, {and} \bibinfo{person}{Chao Xu}.}
  \bibinfo{year}{2015}\natexlab{}.
\newblock \showarticletitle{{Database saliency for fast image retrieval}}.
\newblock \bibinfo{journal}{\emph{IEEE Transactions on Multimedia}}
  \bibinfo{volume}{17}, \bibinfo{number}{3} (\bibinfo{year}{2015}),
  \bibinfo{pages}{359--369}.
\newblock


\bibitem[\protect\citeauthoryear{Guo, Cai, Liu, Mu, Martin, and Hu}{Guo
  et~al\mbox{.}}{2020}]%
        {guo2020pct}
\bibfield{author}{\bibinfo{person}{Meng-Hao Guo}, \bibinfo{person}{Jun-Xiong
  Cai}, \bibinfo{person}{Zheng-Ning Liu}, \bibinfo{person}{Tai-Jiang Mu},
  \bibinfo{person}{Ralph~R Martin}, {and} \bibinfo{person}{Shi-Min Hu}.}
  \bibinfo{year}{2020}\natexlab{}.
\newblock \showarticletitle{{PCT: Point Cloud Transformer}}.
\newblock \bibinfo{journal}{\emph{arXiv preprint arXiv:2012.09688}}
  (\bibinfo{year}{2020}).
\newblock


\bibitem[\protect\citeauthoryear{Han, Xiao, Wu, Guo, Xu, and Wang}{Han
  et~al\mbox{.}}{2021}]%
        {han2021transformer}
\bibfield{author}{\bibinfo{person}{Kai Han}, \bibinfo{person}{An Xiao},
  \bibinfo{person}{Enhua Wu}, \bibinfo{person}{Jianyuan Guo},
  \bibinfo{person}{Chunjing Xu}, {and} \bibinfo{person}{Yunhe Wang}.}
  \bibinfo{year}{2021}\natexlab{}.
\newblock \showarticletitle{{Transformer in transformer}}.
\newblock \bibinfo{journal}{\emph{arXiv preprint arXiv:2103.00112}}
  (\bibinfo{year}{2021}).
\newblock


\bibitem[\protect\citeauthoryear{Hassani, Walton, Shah, Abuduweili, Li, and
  Shi}{Hassani et~al\mbox{.}}{2021}]%
        {hassani2021escaping}
\bibfield{author}{\bibinfo{person}{Ali Hassani}, \bibinfo{person}{Steven
  Walton}, \bibinfo{person}{Nikhil Shah}, \bibinfo{person}{Abulikemu
  Abuduweili}, \bibinfo{person}{Jiachen Li}, {and} \bibinfo{person}{Humphrey
  Shi}.} \bibinfo{year}{2021}\natexlab{}.
\newblock \showarticletitle{{Escaping the Big Data Paradigm with Compact
  Transformers}}.
\newblock \bibinfo{journal}{\emph{arXiv preprint arXiv:2104.05704}}
  (\bibinfo{year}{2021}).
\newblock


\bibitem[\protect\citeauthoryear{He, Zhang, Ren, and Sun}{He
  et~al\mbox{.}}{2016}]%
        {he2016deep}
\bibfield{author}{\bibinfo{person}{Kaiming He}, \bibinfo{person}{Xiangyu
  Zhang}, \bibinfo{person}{Shaoqing Ren}, {and} \bibinfo{person}{Jian Sun}.}
  \bibinfo{year}{2016}\natexlab{}.
\newblock \showarticletitle{{Deep residual learning for image recognition}}. In
  \bibinfo{booktitle}{\emph{Proceedings of the IEEE conference on computer
  vision and pattern recognition}}. \bibinfo{pages}{770--778}.
\newblock


\bibitem[\protect\citeauthoryear{Hong, You, Kwak, and Han}{Hong
  et~al\mbox{.}}{2015}]%
        {hong2015online}
\bibfield{author}{\bibinfo{person}{Seunghoon Hong}, \bibinfo{person}{Tackgeun
  You}, \bibinfo{person}{Suha Kwak}, {and} \bibinfo{person}{Bohyung Han}.}
  \bibinfo{year}{2015}\natexlab{}.
\newblock \showarticletitle{{Online tracking by learning discriminative
  saliency map with convolutional neural network}}. In
  \bibinfo{booktitle}{\emph{International conference on machine learning}}.
  \bibinfo{pages}{597--606}.
\newblock


\bibitem[\protect\citeauthoryear{Huang, Yang, Zhang, Zhang, and Han}{Huang
  et~al\mbox{.}}{2021}]%
        {huang2021employing}
\bibfield{author}{\bibinfo{person}{Nianchang Huang}, \bibinfo{person}{Yang
  Yang}, \bibinfo{person}{Dingwen Zhang}, \bibinfo{person}{Qiang Zhang}, {and}
  \bibinfo{person}{Jungong Han}.} \bibinfo{year}{2021}\natexlab{}.
\newblock \showarticletitle{{Employing Bilinear Fusion and Saliency Prior
  Information for RGB-D Salient Object Detection}}.
\newblock \bibinfo{journal}{\emph{IEEE Transactions on Multimedia}}
  (\bibinfo{year}{2021}).
\newblock


\bibitem[\protect\citeauthoryear{Ji, Fang, Xie, and Lu}{Ji
  et~al\mbox{.}}{2013}]%
        {ji2013video}
\bibfield{author}{\bibinfo{person}{Qing-Ge Ji}, \bibinfo{person}{Zhi-Dang
  Fang}, \bibinfo{person}{Zhen-Hua Xie}, {and} \bibinfo{person}{Zhe-Ming Lu}.}
  \bibinfo{year}{2013}\natexlab{}.
\newblock \showarticletitle{{Video abstraction based on the visual attention
  model and online clustering}}.
\newblock \bibinfo{journal}{\emph{Signal Processing: Image Communication}}
  \bibinfo{volume}{28}, \bibinfo{number}{3} (\bibinfo{year}{2013}),
  \bibinfo{pages}{241--253}.
\newblock


\bibitem[\protect\citeauthoryear{Ji, Li, Zhang, Piao, and Lu}{Ji
  et~al\mbox{.}}{2020}]%
        {ji2020accurate}
\bibfield{author}{\bibinfo{person}{Wei Ji}, \bibinfo{person}{Jingjing Li},
  \bibinfo{person}{Miao Zhang}, \bibinfo{person}{Yongri Piao}, {and}
  \bibinfo{person}{Huchuan Lu}.} \bibinfo{year}{2020}\natexlab{}.
\newblock \showarticletitle{{Accurate rgb-d salient object detection via
  collaborative learning}}. In \bibinfo{booktitle}{\emph{Computer Vision--ECCV
  2020: 16th European Conference, Glasgow, UK, August 23--28, 2020,
  Proceedings, Part XVIII 16}}. Springer, \bibinfo{pages}{52--69}.
\newblock


\bibitem[\protect\citeauthoryear{Jiang, Shao, Lin, Gu, Jiang, and Sun}{Jiang
  et~al\mbox{.}}{2017}]%
        {jiang2017optimizing}
\bibfield{author}{\bibinfo{person}{Qiuping Jiang}, \bibinfo{person}{Feng Shao},
  \bibinfo{person}{Weisi Lin}, \bibinfo{person}{Ke Gu}, \bibinfo{person}{Gangyi
  Jiang}, {and} \bibinfo{person}{Huifang Sun}.}
  \bibinfo{year}{2017}\natexlab{}.
\newblock \showarticletitle{Optimizing multistage discriminative dictionaries
  for blind image quality assessment}.
\newblock \bibinfo{journal}{\emph{IEEE Transactions on Multimedia}}
  \bibinfo{volume}{20}, \bibinfo{number}{8} (\bibinfo{year}{2017}),
  \bibinfo{pages}{2035--2048}.
\newblock


\bibitem[\protect\citeauthoryear{Jin, Xu, Han, Zhang, and Cheng}{Jin
  et~al\mbox{.}}{2021}]%
        {jin2021cdnet}
\bibfield{author}{\bibinfo{person}{Wen-Da Jin}, \bibinfo{person}{Jun Xu},
  \bibinfo{person}{Qi Han}, \bibinfo{person}{Yi Zhang}, {and}
  \bibinfo{person}{Ming-Ming Cheng}.} \bibinfo{year}{2021}\natexlab{}.
\newblock \showarticletitle{{CDNet: Complementary Depth Network for RGB-D
  Salient Object Detection}}.
\newblock \bibinfo{journal}{\emph{IEEE Transactions on Image Processing}}
  \bibinfo{volume}{30} (\bibinfo{year}{2021}), \bibinfo{pages}{3376--3390}.
\newblock


\bibitem[\protect\citeauthoryear{Ju, Ge, Geng, Ren, and Wu}{Ju
  et~al\mbox{.}}{2014}]%
        {ju2014depth}
\bibfield{author}{\bibinfo{person}{Ran Ju}, \bibinfo{person}{Ling Ge},
  \bibinfo{person}{Wenjing Geng}, \bibinfo{person}{Tongwei Ren}, {and}
  \bibinfo{person}{Gangshan Wu}.} \bibinfo{year}{2014}\natexlab{}.
\newblock \showarticletitle{{Depth saliency based on anisotropic
  center-surround difference}}. In \bibinfo{booktitle}{\emph{2014 IEEE
  international conference on image processing (ICIP)}}. IEEE,
  \bibinfo{pages}{1115--1119}.
\newblock


\bibitem[\protect\citeauthoryear{Kingma and Ba}{Kingma and Ba}{2014}]%
        {kingma2014adam}
\bibfield{author}{\bibinfo{person}{Diederik~P Kingma} {and}
  \bibinfo{person}{Jimmy Ba}.} \bibinfo{year}{2014}\natexlab{}.
\newblock \showarticletitle{{Adam: A method for stochastic optimization}}.
\newblock \bibinfo{journal}{\emph{arXiv preprint arXiv:1412.6980}}
  (\bibinfo{year}{2014}).
\newblock


\bibitem[\protect\citeauthoryear{Li, Cong, Kwong, Hou, Fu, Zhu, Zhang, and
  Huang}{Li et~al\mbox{.}}{2020a}]%
        {li2020asif}
\bibfield{author}{\bibinfo{person}{Chongyi Li}, \bibinfo{person}{Runmin Cong},
  \bibinfo{person}{Sam Kwong}, \bibinfo{person}{Junhui Hou},
  \bibinfo{person}{Huazhu Fu}, \bibinfo{person}{Guopu Zhu},
  \bibinfo{person}{Dingwen Zhang}, {and} \bibinfo{person}{Qingming Huang}.}
  \bibinfo{year}{2020}\natexlab{a}.
\newblock \showarticletitle{{ASIF-Net: Attention steered interweave fusion
  network for RGB-D salient object detection}}.
\newblock \bibinfo{journal}{\emph{IEEE Transactions on Cybernetics}}
  (\bibinfo{year}{2020}).
\newblock


\bibitem[\protect\citeauthoryear{Li, Cong, Piao, Xu, and Loy}{Li
  et~al\mbox{.}}{2020b}]%
        {li2020rgb}
\bibfield{author}{\bibinfo{person}{Chongyi Li}, \bibinfo{person}{Runmin Cong},
  \bibinfo{person}{Yongri Piao}, \bibinfo{person}{Qianqian Xu}, {and}
  \bibinfo{person}{Chen~Change Loy}.} \bibinfo{year}{2020}\natexlab{b}.
\newblock \showarticletitle{{RGB-D salient object detection with cross-modality
  modulation and selection}}. In \bibinfo{booktitle}{\emph{European Conference
  on Computer Vision}}. Springer, \bibinfo{pages}{225--241}.
\newblock


\bibitem[\protect\citeauthoryear{Li, Liu, Chen, Bai, Lin, and Ling}{Li
  et~al\mbox{.}}{2021a}]%
        {li2021hierarchical}
\bibfield{author}{\bibinfo{person}{Gongyang Li}, \bibinfo{person}{Zhi Liu},
  \bibinfo{person}{Minyu Chen}, \bibinfo{person}{Zhen Bai},
  \bibinfo{person}{Weisi Lin}, {and} \bibinfo{person}{Haibin Ling}.}
  \bibinfo{year}{2021}\natexlab{a}.
\newblock \showarticletitle{Hierarchical Alternate Interaction Network for
  RGB-D Salient Object Detection}.
\newblock \bibinfo{journal}{\emph{IEEE Transactions on Image Processing}}
  \bibinfo{volume}{30} (\bibinfo{year}{2021}), \bibinfo{pages}{3528--3542}.
\newblock


\bibitem[\protect\citeauthoryear{Li, Liu, and Ling}{Li et~al\mbox{.}}{2020c}]%
        {li2020icnet}
\bibfield{author}{\bibinfo{person}{Gongyang Li}, \bibinfo{person}{Zhi Liu},
  {and} \bibinfo{person}{Haibin Ling}.} \bibinfo{year}{2020}\natexlab{c}.
\newblock \showarticletitle{{ICNet: Information Conversion Network for RGB-D
  Based Salient Object Detection}}.
\newblock \bibinfo{journal}{\emph{IEEE Transactions on Image Processing}}
  \bibinfo{volume}{29} (\bibinfo{year}{2020}), \bibinfo{pages}{4873--4884}.
\newblock


\bibitem[\protect\citeauthoryear{Li, Zhang, Cao, Timofte, and Van~Gool}{Li
  et~al\mbox{.}}{2021b}]%
        {li2021localvit}
\bibfield{author}{\bibinfo{person}{Yawei Li}, \bibinfo{person}{Kai Zhang},
  \bibinfo{person}{Jiezhang Cao}, \bibinfo{person}{Radu Timofte}, {and}
  \bibinfo{person}{Luc Van~Gool}.} \bibinfo{year}{2021}\natexlab{b}.
\newblock \showarticletitle{{LocalViT: Bringing Locality to Vision
  Transformers}}.
\newblock \bibinfo{journal}{\emph{arXiv preprint arXiv:2104.05707}}
  (\bibinfo{year}{2021}).
\newblock


\bibitem[\protect\citeauthoryear{Liu, Zhang, and Han}{Liu
  et~al\mbox{.}}{2020a}]%
        {liu2020learning}
\bibfield{author}{\bibinfo{person}{Nian Liu}, \bibinfo{person}{Ni Zhang}, {and}
  \bibinfo{person}{Junwei Han}.} \bibinfo{year}{2020}\natexlab{a}.
\newblock \showarticletitle{{Learning Selective Self-Mutual Attention for RGB-D
  Saliency Detection}}. In \bibinfo{booktitle}{\emph{Proceedings of the
  IEEE/CVF Conference on Computer Vision and Pattern Recognition}}.
  \bibinfo{pages}{13756--13765}.
\newblock


\bibitem[\protect\citeauthoryear{Liu, Lin, Cao, Hu, Wei, Zhang, Lin, and
  Guo}{Liu et~al\mbox{.}}{2021}]%
        {liu2021swin}
\bibfield{author}{\bibinfo{person}{Ze Liu}, \bibinfo{person}{Yutong Lin},
  \bibinfo{person}{Yue Cao}, \bibinfo{person}{Han Hu}, \bibinfo{person}{Yixuan
  Wei}, \bibinfo{person}{Zheng Zhang}, \bibinfo{person}{Stephen Lin}, {and}
  \bibinfo{person}{Baining Guo}.} \bibinfo{year}{2021}\natexlab{}.
\newblock \showarticletitle{{Swin transformer: Hierarchical vision transformer
  using shifted windows}}.
\newblock \bibinfo{journal}{\emph{arXiv preprint arXiv:2103.14030}}
  (\bibinfo{year}{2021}).
\newblock


\bibitem[\protect\citeauthoryear{Liu, Shi, Duan, Zhang, and Zhao}{Liu
  et~al\mbox{.}}{2019}]%
        {liu2019salient}
\bibfield{author}{\bibinfo{person}{Zhengyi Liu}, \bibinfo{person}{Song Shi},
  \bibinfo{person}{Quntao Duan}, \bibinfo{person}{Wei Zhang}, {and}
  \bibinfo{person}{Peng Zhao}.} \bibinfo{year}{2019}\natexlab{}.
\newblock \showarticletitle{{Salient object detection for RGB-D image by single
  stream recurrent convolution neural network}}.
\newblock \bibinfo{journal}{\emph{Neurocomputing}}  \bibinfo{volume}{363}
  (\bibinfo{year}{2019}), \bibinfo{pages}{46--57}.
\newblock


\bibitem[\protect\citeauthoryear{Liu, Zhang, and Zhao}{Liu
  et~al\mbox{.}}{2020b}]%
        {liu2020cross}
\bibfield{author}{\bibinfo{person}{Zhengyi Liu}, \bibinfo{person}{Wei Zhang},
  {and} \bibinfo{person}{Peng Zhao}.} \bibinfo{year}{2020}\natexlab{b}.
\newblock \showarticletitle{{A cross-modal adaptive gated fusion generative
  adversarial network for RGB-D salient object detection}}.
\newblock \bibinfo{journal}{\emph{Neurocomputing}}  \bibinfo{volume}{387}
  (\bibinfo{year}{2020}), \bibinfo{pages}{210--220}.
\newblock


\bibitem[\protect\citeauthoryear{Ma, Miao, Zhang, and Li}{Ma
  et~al\mbox{.}}{2017}]%
        {ma2017saliency}
\bibfield{author}{\bibinfo{person}{Cong Ma}, \bibinfo{person}{Zhenjiang Miao},
  \bibinfo{person}{Xiao-Ping Zhang}, {and} \bibinfo{person}{Min Li}.}
  \bibinfo{year}{2017}\natexlab{}.
\newblock \showarticletitle{{A saliency prior context model for real-time
  object tracking}}.
\newblock \bibinfo{journal}{\emph{IEEE Transactions on Multimedia}}
  \bibinfo{volume}{19}, \bibinfo{number}{11} (\bibinfo{year}{2017}),
  \bibinfo{pages}{2415--2424}.
\newblock


\bibitem[\protect\citeauthoryear{Ma, Sun, and Li}{Ma et~al\mbox{.}}{2021}]%
        {ma2021robust}
\bibfield{author}{\bibinfo{person}{Fuyan Ma}, \bibinfo{person}{Bin Sun}, {and}
  \bibinfo{person}{Shutao Li}.} \bibinfo{year}{2021}\natexlab{}.
\newblock \showarticletitle{{Robust Facial Expression Recognition with
  Convolutional Visual Transformers}}.
\newblock \bibinfo{journal}{\emph{arXiv preprint arXiv:2103.16854}}
  (\bibinfo{year}{2021}).
\newblock


\bibitem[\protect\citeauthoryear{Meinhardt, Kirillov, Leal-Taixe, and
  Feichtenhofer}{Meinhardt et~al\mbox{.}}{2021}]%
        {meinhardt2021trackformer}
\bibfield{author}{\bibinfo{person}{Tim Meinhardt}, \bibinfo{person}{Alexander
  Kirillov}, \bibinfo{person}{Laura Leal-Taixe}, {and}
  \bibinfo{person}{Christoph Feichtenhofer}.} \bibinfo{year}{2021}\natexlab{}.
\newblock \showarticletitle{TrackFormer: Multi-Object Tracking with
  Transformers}.
\newblock \bibinfo{journal}{\emph{arXiv preprint arXiv:2101.02702}}
  (\bibinfo{year}{2021}).
\newblock


\bibitem[\protect\citeauthoryear{Niu, Geng, Li, and Liu}{Niu
  et~al\mbox{.}}{2012}]%
        {niu2012leveraging}
\bibfield{author}{\bibinfo{person}{Yuzhen Niu}, \bibinfo{person}{Yujie Geng},
  \bibinfo{person}{Xueqing Li}, {and} \bibinfo{person}{Feng Liu}.}
  \bibinfo{year}{2012}\natexlab{}.
\newblock \showarticletitle{{Leveraging stereopsis for saliency analysis}}. In
  \bibinfo{booktitle}{\emph{2012 IEEE Conference on Computer Vision and Pattern
  Recognition}}. IEEE, \bibinfo{pages}{454--461}.
\newblock


\bibitem[\protect\citeauthoryear{Ouerhani and Hugli}{Ouerhani and
  Hugli}{2000}]%
        {ouerhani2000computing}
\bibfield{author}{\bibinfo{person}{Nabil Ouerhani} {and} \bibinfo{person}{Heinz
  Hugli}.} \bibinfo{year}{2000}\natexlab{}.
\newblock \showarticletitle{{Computing visual attention from scene depth}}. In
  \bibinfo{booktitle}{\emph{Proceedings 15th International Conference on
  Pattern Recognition. ICPR-2000}}, Vol.~\bibinfo{volume}{1}. IEEE,
  \bibinfo{pages}{375--378}.
\newblock


\bibitem[\protect\citeauthoryear{Pan, Zhou, Shi, Zhang, and Yan}{Pan
  et~al\mbox{.}}{2020}]%
        {pan2020cross}
\bibfield{author}{\bibinfo{person}{Liang Pan}, \bibinfo{person}{Xiaofei Zhou},
  \bibinfo{person}{Ran Shi}, \bibinfo{person}{Jiyong Zhang}, {and}
  \bibinfo{person}{Chenggang Yan}.} \bibinfo{year}{2020}\natexlab{}.
\newblock \showarticletitle{{Cross-modal feature extraction and integration
  based RGBD saliency detection}}.
\newblock \bibinfo{journal}{\emph{Image and Vision Computing}}
  \bibinfo{volume}{101} (\bibinfo{year}{2020}), \bibinfo{pages}{103964}.
\newblock


\bibitem[\protect\citeauthoryear{Pang, Zhang, Zhao, and Lu}{Pang
  et~al\mbox{.}}{2020}]%
        {pang2020hierarchical}
\bibfield{author}{\bibinfo{person}{Youwei Pang}, \bibinfo{person}{Lihe Zhang},
  \bibinfo{person}{Xiaoqi Zhao}, {and} \bibinfo{person}{Huchuan Lu}.}
  \bibinfo{year}{2020}\natexlab{}.
\newblock \showarticletitle{Hierarchical dynamic filtering network for rgb-d
  salient object detection}. In \bibinfo{booktitle}{\emph{Computer Vision--ECCV
  2020: 16th European Conference, Glasgow, UK, August 23--28, 2020,
  Proceedings, Part XXV 16}}. Springer, \bibinfo{pages}{235--252}.
\newblock


\bibitem[\protect\citeauthoryear{Peng, Li, Xiong, Hu, and Ji}{Peng
  et~al\mbox{.}}{2014}]%
        {peng2014rgbd}
\bibfield{author}{\bibinfo{person}{Houwen Peng}, \bibinfo{person}{Bing Li},
  \bibinfo{person}{Weihua Xiong}, \bibinfo{person}{Weiming Hu}, {and}
  \bibinfo{person}{Rongrong Ji}.} \bibinfo{year}{2014}\natexlab{}.
\newblock \showarticletitle{{Rgbd salient object detection: a benchmark and
  algorithms}}. In \bibinfo{booktitle}{\emph{European conference on computer
  vision}}. Springer, \bibinfo{pages}{92--109}.
\newblock


\bibitem[\protect\citeauthoryear{Perazzi, Kr{\"a}henb{\"u}hl, Pritch, and
  Hornung}{Perazzi et~al\mbox{.}}{2012}]%
        {perazzi2012saliency}
\bibfield{author}{\bibinfo{person}{Federico Perazzi}, \bibinfo{person}{Philipp
  Kr{\"a}henb{\"u}hl}, \bibinfo{person}{Yael Pritch}, {and}
  \bibinfo{person}{Alexander Hornung}.} \bibinfo{year}{2012}\natexlab{}.
\newblock \showarticletitle{{Saliency filters: Contrast based filtering for
  salient region detection}}. In \bibinfo{booktitle}{\emph{2012 IEEE conference
  on computer vision and pattern recognition}}. IEEE,
  \bibinfo{pages}{733--740}.
\newblock


\bibitem[\protect\citeauthoryear{Piao, Ji, Li, Zhang, and Lu}{Piao
  et~al\mbox{.}}{2019}]%
        {piao2019depth}
\bibfield{author}{\bibinfo{person}{Yongri Piao}, \bibinfo{person}{Wei Ji},
  \bibinfo{person}{Jingjing Li}, \bibinfo{person}{Miao Zhang}, {and}
  \bibinfo{person}{Huchuan Lu}.} \bibinfo{year}{2019}\natexlab{}.
\newblock \showarticletitle{{Depth-induced multi-scale recurrent attention
  network for saliency detection}}. In \bibinfo{booktitle}{\emph{Proceedings of
  the IEEE International Conference on Computer Vision}}.
  \bibinfo{pages}{7254--7263}.
\newblock


\bibitem[\protect\citeauthoryear{Piao, Rong, Zhang, Ren, and Lu}{Piao
  et~al\mbox{.}}{2020}]%
        {piao2020a2dele}
\bibfield{author}{\bibinfo{person}{Yongri Piao}, \bibinfo{person}{Zhengkun
  Rong}, \bibinfo{person}{Miao Zhang}, \bibinfo{person}{Weisong Ren}, {and}
  \bibinfo{person}{Huchuan Lu}.} \bibinfo{year}{2020}\natexlab{}.
\newblock \showarticletitle{{A2dele: Adaptive and Attentive Depth Distiller for
  Efficient RGB-D Salient Object Detection}}. In
  \bibinfo{booktitle}{\emph{Proceedings of the IEEE/CVF Conference on Computer
  Vision and Pattern Recognition}}. \bibinfo{pages}{9060--9069}.
\newblock


\bibitem[\protect\citeauthoryear{Ronneberger, Fischer, and Brox}{Ronneberger
  et~al\mbox{.}}{2015}]%
        {ronneberger2015u}
\bibfield{author}{\bibinfo{person}{Olaf Ronneberger}, \bibinfo{person}{Philipp
  Fischer}, {and} \bibinfo{person}{Thomas Brox}.}
  \bibinfo{year}{2015}\natexlab{}.
\newblock \showarticletitle{{U-Net: Convolutional networks for biomedical image
  segmentation}}. In \bibinfo{booktitle}{\emph{International Conference on
  Medical image computing and computer-assisted intervention}}. Springer,
  \bibinfo{pages}{234--241}.
\newblock


\bibitem[\protect\citeauthoryear{Stoffl, Vidal, and Mathis}{Stoffl
  et~al\mbox{.}}{2021}]%
        {stoffl2021end}
\bibfield{author}{\bibinfo{person}{Lucas Stoffl}, \bibinfo{person}{Maxime
  Vidal}, {and} \bibinfo{person}{Alexander Mathis}.}
  \bibinfo{year}{2021}\natexlab{}.
\newblock \showarticletitle{{End-to-End Trainable Multi-Instance Pose
  Estimation with Transformers}}.
\newblock \bibinfo{journal}{\emph{arXiv preprint arXiv:2103.12115}}
  (\bibinfo{year}{2021}).
\newblock


\bibitem[\protect\citeauthoryear{Sun, Yang, Hu, Hu, and Wang}{Sun
  et~al\mbox{.}}{2020}]%
        {sun2020real}
\bibfield{author}{\bibinfo{person}{Lei Sun}, \bibinfo{person}{Kailun Yang},
  \bibinfo{person}{Xinxin Hu}, \bibinfo{person}{Weijian Hu}, {and}
  \bibinfo{person}{Kaiwei Wang}.} \bibinfo{year}{2020}\natexlab{}.
\newblock \showarticletitle{{Real-time fusion network for RGB-D semantic
  segmentation incorporating unexpected obstacle detection for road-driving
  images}}.
\newblock \bibinfo{journal}{\emph{IEEE Robotics and Automation Letters}}
  \bibinfo{volume}{5}, \bibinfo{number}{4} (\bibinfo{year}{2020}),
  \bibinfo{pages}{5558--5565}.
\newblock


\bibitem[\protect\citeauthoryear{Sun, Zhang, Wang, Li, and Li}{Sun
  et~al\mbox{.}}{2021}]%
        {Sun2021DeepRS}
\bibfield{author}{\bibinfo{person}{Peng Sun}, \bibinfo{person}{Wenhu Zhang},
  \bibinfo{person}{Huanyu Wang}, \bibinfo{person}{Songyuan Li}, {and}
  \bibinfo{person}{Xi Li}.} \bibinfo{year}{2021}\natexlab{}.
\newblock \showarticletitle{{Deep RGB-D Saliency Detection with Depth-Sensitive
  Attention and Automatic Multi-Modal Fusion}}. In
  \bibinfo{booktitle}{\emph{Proceedings of the IEEE/CVF Conference on Computer
  Vision and Pattern Recognition}}. \bibinfo{pages}{1407--1417}.
\newblock


\bibitem[\protect\citeauthoryear{Vaswani, Shazeer, Parmar, Uszkoreit, Jones,
  Gomez, Kaiser, and Polosukhin}{Vaswani et~al\mbox{.}}{2017}]%
        {NIPS2017_3f5ee243}
\bibfield{author}{\bibinfo{person}{Ashish Vaswani}, \bibinfo{person}{Noam
  Shazeer}, \bibinfo{person}{Niki Parmar}, \bibinfo{person}{Jakob Uszkoreit},
  \bibinfo{person}{Llion Jones}, \bibinfo{person}{Aidan~N Gomez},
  \bibinfo{person}{\L~ukasz Kaiser}, {and} \bibinfo{person}{Illia Polosukhin}.}
  \bibinfo{year}{2017}\natexlab{}.
\newblock \showarticletitle{Attention is All you Need}. In
  \bibinfo{booktitle}{\emph{Advances in Neural Information Processing
  Systems}}, Vol.~\bibinfo{volume}{30}. \bibinfo{publisher}{Curran Associates,
  Inc.}, \bibinfo{pages}{5998--6008}.
\newblock


\bibitem[\protect\citeauthoryear{Wan, Zhang, Chen, and Liao}{Wan
  et~al\mbox{.}}{2021}]%
        {wan2021high}
\bibfield{author}{\bibinfo{person}{Ziyu Wan}, \bibinfo{person}{Jingbo Zhang},
  \bibinfo{person}{Dongdong Chen}, {and} \bibinfo{person}{Jing Liao}.}
  \bibinfo{year}{2021}\natexlab{}.
\newblock \showarticletitle{{High-Fidelity Pluralistic Image Completion with
  Transformers}}.
\newblock \bibinfo{journal}{\emph{arXiv preprint arXiv:2103.14031}}
  (\bibinfo{year}{2021}).
\newblock


\bibitem[\protect\citeauthoryear{Wang and Gong}{Wang and Gong}{2019}]%
        {wang2019adaptive}
\bibfield{author}{\bibinfo{person}{Ningning Wang} {and}
  \bibinfo{person}{Xiaojin Gong}.} \bibinfo{year}{2019}\natexlab{}.
\newblock \showarticletitle{{Adaptive fusion for RGB-D salient object
  detection}}.
\newblock \bibinfo{journal}{\emph{IEEE Access}}  \bibinfo{volume}{7}
  (\bibinfo{year}{2019}), \bibinfo{pages}{55277--55284}.
\newblock


\bibitem[\protect\citeauthoryear{Wang, Shen, and Ling}{Wang
  et~al\mbox{.}}{2018}]%
        {wang2018deep}
\bibfield{author}{\bibinfo{person}{Wenguan Wang}, \bibinfo{person}{Jianbing
  Shen}, {and} \bibinfo{person}{Haibin Ling}.} \bibinfo{year}{2018}\natexlab{}.
\newblock \showarticletitle{{A deep network solution for attention and
  aesthetics aware photo cropping}}.
\newblock \bibinfo{journal}{\emph{IEEE transactions on pattern analysis and
  machine intelligence}} \bibinfo{volume}{41}, \bibinfo{number}{7}
  (\bibinfo{year}{2018}), \bibinfo{pages}{1531--1544}.
\newblock


\bibitem[\protect\citeauthoryear{Wang, Xie, Li, Fan, Song, Liang, Lu, Luo, and
  Shao}{Wang et~al\mbox{.}}{2021}]%
        {wang2021pyramid}
\bibfield{author}{\bibinfo{person}{Wenhai Wang}, \bibinfo{person}{Enze Xie},
  \bibinfo{person}{Xiang Li}, \bibinfo{person}{Deng-Ping Fan},
  \bibinfo{person}{Kaitao Song}, \bibinfo{person}{Ding Liang},
  \bibinfo{person}{Tong Lu}, \bibinfo{person}{Ping Luo}, {and}
  \bibinfo{person}{Ling Shao}.} \bibinfo{year}{2021}\natexlab{}.
\newblock \showarticletitle{Pyramid vision transformer: A versatile backbone
  for dense prediction without convolutions}.
\newblock \bibinfo{journal}{\emph{arXiv preprint arXiv:2102.12122}}
  (\bibinfo{year}{2021}).
\newblock


\bibitem[\protect\citeauthoryear{Wang, Li, Chen, Fang, Hao, and Qin}{Wang
  et~al\mbox{.}}{2020a}]%
        {wang2020data}
\bibfield{author}{\bibinfo{person}{Xuehao Wang}, \bibinfo{person}{Shuai Li},
  \bibinfo{person}{Chenglizhao Chen}, \bibinfo{person}{Yuming Fang},
  \bibinfo{person}{Aimin Hao}, {and} \bibinfo{person}{Hong Qin}.}
  \bibinfo{year}{2020}\natexlab{a}.
\newblock \showarticletitle{{Data-level recombination and lightweight fusion
  scheme for RGB-D salient object detection}}.
\newblock \bibinfo{journal}{\emph{IEEE Transactions on Image Processing}}
  \bibinfo{volume}{30} (\bibinfo{year}{2020}), \bibinfo{pages}{458--471}.
\newblock


\bibitem[\protect\citeauthoryear{Wang, Li, Elder, Wu, Lu, and Zhang}{Wang
  et~al\mbox{.}}{2020b}]%
        {wang2020synergistic}
\bibfield{author}{\bibinfo{person}{Yue Wang}, \bibinfo{person}{Yuke Li},
  \bibinfo{person}{James~H Elder}, \bibinfo{person}{Runmin Wu},
  \bibinfo{person}{Huchuan Lu}, {and} \bibinfo{person}{Lu Zhang}.}
  \bibinfo{year}{2020}\natexlab{b}.
\newblock \showarticletitle{{Synergistic saliency and depth prediction for
  RGB-D saliency detection}}. In \bibinfo{booktitle}{\emph{Proceedings of the
  Asian Conference on Computer Vision}}. \bibinfo{pages}{1--17}.
\newblock


\bibitem[\protect\citeauthoryear{Wei, Wang, and Huang}{Wei
  et~al\mbox{.}}{2020}]%
        {wei2020f3net}
\bibfield{author}{\bibinfo{person}{Jun Wei}, \bibinfo{person}{Shuhui Wang},
  {and} \bibinfo{person}{Qingming Huang}.} \bibinfo{year}{2020}\natexlab{}.
\newblock \showarticletitle{{F$^3$Net: Fusion, Feedback and Focus for Salient
  Object Detection}}. In \bibinfo{booktitle}{\emph{Proceedings of the AAAI
  Conference on Artificial Intelligence}}. \bibinfo{pages}{12321--12328}.
\newblock


\bibitem[\protect\citeauthoryear{Woo, Park, Lee, and So~Kweon}{Woo
  et~al\mbox{.}}{2018}]%
        {woo2018cbam}
\bibfield{author}{\bibinfo{person}{Sanghyun Woo}, \bibinfo{person}{Jongchan
  Park}, \bibinfo{person}{Joon-Young Lee}, {and} \bibinfo{person}{In
  So~Kweon}.} \bibinfo{year}{2018}\natexlab{}.
\newblock \showarticletitle{{CBAM: Convolutional block attention module}}. In
  \bibinfo{booktitle}{\emph{Proceedings of the European conference on computer
  vision (ECCV)}}. \bibinfo{pages}{3--19}.
\newblock


\bibitem[\protect\citeauthoryear{Wu, Xu, Dai, Wan, Zhang, Tomizuka, Keutzer,
  and Vajda}{Wu et~al\mbox{.}}{2020b}]%
        {wu2020visual}
\bibfield{author}{\bibinfo{person}{Bichen Wu}, \bibinfo{person}{Chenfeng Xu},
  \bibinfo{person}{Xiaoliang Dai}, \bibinfo{person}{Alvin Wan},
  \bibinfo{person}{Peizhao Zhang}, \bibinfo{person}{Masayoshi Tomizuka},
  \bibinfo{person}{Kurt Keutzer}, {and} \bibinfo{person}{Peter Vajda}.}
  \bibinfo{year}{2020}\natexlab{b}.
\newblock \showarticletitle{Visual transformers: Token-based image
  representation and processing for computer vision}.
\newblock \bibinfo{journal}{\emph{arXiv preprint arXiv:2006.03677}}
  (\bibinfo{year}{2020}).
\newblock


\bibitem[\protect\citeauthoryear{Wu, Xiao, Codella, Liu, Dai, Yuan, and
  Zhang}{Wu et~al\mbox{.}}{2021}]%
        {wu2021cvt}
\bibfield{author}{\bibinfo{person}{Haiping Wu}, \bibinfo{person}{Bin Xiao},
  \bibinfo{person}{Noel Codella}, \bibinfo{person}{Mengchen Liu},
  \bibinfo{person}{Xiyang Dai}, \bibinfo{person}{Lu Yuan}, {and}
  \bibinfo{person}{Lei Zhang}.} \bibinfo{year}{2021}\natexlab{}.
\newblock \showarticletitle{CvT: Introducing Convolutions to Vision
  Transformers}.
\newblock \bibinfo{journal}{\emph{arXiv preprint arXiv:2103.15808}}
  (\bibinfo{year}{2021}).
\newblock


\bibitem[\protect\citeauthoryear{Wu, Liu, Xu, Bian, Gu, and Cheng}{Wu
  et~al\mbox{.}}{2020a}]%
        {wu2020mobilesal}
\bibfield{author}{\bibinfo{person}{Yu-Huan Wu}, \bibinfo{person}{Yun Liu},
  \bibinfo{person}{Jun Xu}, \bibinfo{person}{Jia-Wang Bian},
  \bibinfo{person}{Yuchao Gu}, {and} \bibinfo{person}{Ming-Ming Cheng}.}
  \bibinfo{year}{2020}\natexlab{a}.
\newblock \showarticletitle{{MobileSal: Extremely Efficient RGB-D Salient
  Object Detection}}.
\newblock \bibinfo{journal}{\emph{arXiv preprint arXiv:2012.13095}}
  (\bibinfo{year}{2020}).
\newblock


\bibitem[\protect\citeauthoryear{Wu, Su, and Huang}{Wu et~al\mbox{.}}{2019}]%
        {wu2019cascaded}
\bibfield{author}{\bibinfo{person}{Zhe Wu}, \bibinfo{person}{Li Su}, {and}
  \bibinfo{person}{Qingming Huang}.} \bibinfo{year}{2019}\natexlab{}.
\newblock \showarticletitle{{Cascaded partial decoder for fast and accurate
  salient object detection}}. In \bibinfo{booktitle}{\emph{Proceedings of the
  IEEE Conference on Computer Vision and Pattern Recognition}}.
  \bibinfo{pages}{3907--3916}.
\newblock


\bibitem[\protect\citeauthoryear{Xie, Zhang, Shen, and Xia}{Xie
  et~al\mbox{.}}{2021}]%
        {xie2021cotr}
\bibfield{author}{\bibinfo{person}{Yutong Xie}, \bibinfo{person}{Jianpeng
  Zhang}, \bibinfo{person}{Chunhua Shen}, {and} \bibinfo{person}{Yong Xia}.}
  \bibinfo{year}{2021}\natexlab{}.
\newblock \showarticletitle{{CoTr: Efficiently Bridging CNN and Transformer for
  3D Medical Image Segmentation}}.
\newblock \bibinfo{journal}{\emph{arXiv preprint arXiv:2103.03024}}
  (\bibinfo{year}{2021}).
\newblock


\bibitem[\protect\citeauthoryear{Xu, Xu, Chang, and Tu}{Xu
  et~al\mbox{.}}{2021a}]%
        {xu2021co}
\bibfield{author}{\bibinfo{person}{Weijian Xu}, \bibinfo{person}{Yifan Xu},
  \bibinfo{person}{Tyler Chang}, {and} \bibinfo{person}{Zhuowen Tu}.}
  \bibinfo{year}{2021}\natexlab{a}.
\newblock \showarticletitle{{Co-Scale Conv-Attentional Image Transformers}}.
\newblock \bibinfo{journal}{\emph{arXiv preprint arXiv:2104.06399}}
  (\bibinfo{year}{2021}).
\newblock


\bibitem[\protect\citeauthoryear{Xu, Xu, Cheung, and Tu}{Xu
  et~al\mbox{.}}{2021b}]%
        {xu2021line}
\bibfield{author}{\bibinfo{person}{Yifan Xu}, \bibinfo{person}{Weijian Xu},
  \bibinfo{person}{David Cheung}, {and} \bibinfo{person}{Zhuowen Tu}.}
  \bibinfo{year}{2021}\natexlab{b}.
\newblock \showarticletitle{{Line segment detection using transformers without
  edges}}. In \bibinfo{booktitle}{\emph{Proceedings of the IEEE/CVF Conference
  on Computer Vision and Pattern Recognition}}. \bibinfo{pages}{4257--4266}.
\newblock


\bibitem[\protect\citeauthoryear{Yuan, Chen, Wang, Yu, Shi, Tay, Feng, and
  Yan}{Yuan et~al\mbox{.}}{2021}]%
        {yuan2021tokens}
\bibfield{author}{\bibinfo{person}{Li Yuan}, \bibinfo{person}{Yunpeng Chen},
  \bibinfo{person}{Tao Wang}, \bibinfo{person}{Weihao Yu},
  \bibinfo{person}{Yujun Shi}, \bibinfo{person}{Francis~EH Tay},
  \bibinfo{person}{Jiashi Feng}, {and} \bibinfo{person}{Shuicheng Yan}.}
  \bibinfo{year}{2021}\natexlab{}.
\newblock \showarticletitle{{Tokens-to-token vit: Training vision transformers
  from scratch on imagenet}}.
\newblock \bibinfo{journal}{\emph{arXiv preprint arXiv:2101.11986}}
  (\bibinfo{year}{2021}).
\newblock


\bibitem[\protect\citeauthoryear{Zeng, Tong, Huang, Yan, Sun, Chen, and
  Wang}{Zeng et~al\mbox{.}}{2019}]%
        {zeng2019deep}
\bibfield{author}{\bibinfo{person}{Jin Zeng}, \bibinfo{person}{Yanfeng Tong},
  \bibinfo{person}{Yunmu Huang}, \bibinfo{person}{Qiong Yan},
  \bibinfo{person}{Wenxiu Sun}, \bibinfo{person}{Jing Chen}, {and}
  \bibinfo{person}{Yongtian Wang}.} \bibinfo{year}{2019}\natexlab{}.
\newblock \showarticletitle{{Deep surface normal estimation with hierarchical
  rgb-d fusion}}. In \bibinfo{booktitle}{\emph{Proceedings of the IEEE/CVF
  Conference on Computer Vision and Pattern Recognition}}.
  \bibinfo{pages}{6153--6162}.
\newblock


\bibitem[\protect\citeauthoryear{Zhang, Fan, Dai, Anwar, Saleh, Zhang, and
  Barnes}{Zhang et~al\mbox{.}}{2020a}]%
        {zhang2020uc}
\bibfield{author}{\bibinfo{person}{Jing Zhang}, \bibinfo{person}{Deng-Ping
  Fan}, \bibinfo{person}{Yuchao Dai}, \bibinfo{person}{Saeed Anwar},
  \bibinfo{person}{Fatemeh~Sadat Saleh}, \bibinfo{person}{Tong Zhang}, {and}
  \bibinfo{person}{Nick Barnes}.} \bibinfo{year}{2020}\natexlab{a}.
\newblock \showarticletitle{{UC-Net: uncertainty inspired rgb-d saliency
  detection via conditional variational autoencoders}}. In
  \bibinfo{booktitle}{\emph{Proceedings of the IEEE/CVF Conference on Computer
  Vision and Pattern Recognition}}. \bibinfo{pages}{8582--8591}.
\newblock


\bibitem[\protect\citeauthoryear{Zhang, Ren, Piao, Rong, and Lu}{Zhang
  et~al\mbox{.}}{2020c}]%
        {zhang2020select}
\bibfield{author}{\bibinfo{person}{Miao Zhang}, \bibinfo{person}{Weisong Ren},
  \bibinfo{person}{Yongri Piao}, \bibinfo{person}{Zhengkun Rong}, {and}
  \bibinfo{person}{Huchuan Lu}.} \bibinfo{year}{2020}\natexlab{c}.
\newblock \showarticletitle{{Select, Supplement and Focus for RGB-D Saliency
  Detection}}. In \bibinfo{booktitle}{\emph{Proceedings of the IEEE/CVF
  Conference on Computer Vision and Pattern Recognition}}.
  \bibinfo{pages}{3472--3481}.
\newblock


\bibitem[\protect\citeauthoryear{Zhang, Dai, Yang, Xiao, Yuan, Zhang, and
  Gao}{Zhang et~al\mbox{.}}{2021a}]%
        {zhang2021multi}
\bibfield{author}{\bibinfo{person}{Pengchuan Zhang}, \bibinfo{person}{Xiyang
  Dai}, \bibinfo{person}{Jianwei Yang}, \bibinfo{person}{Bin Xiao},
  \bibinfo{person}{Lu Yuan}, \bibinfo{person}{Lei Zhang}, {and}
  \bibinfo{person}{Jianfeng Gao}.} \bibinfo{year}{2021}\natexlab{a}.
\newblock \showarticletitle{{Multi-Scale Vision Longformer: A New Vision
  Transformer for High-Resolution Image Encoding}}.
\newblock \bibinfo{journal}{\emph{arXiv preprint arXiv:2103.15358}}
  (\bibinfo{year}{2021}).
\newblock


\bibitem[\protect\citeauthoryear{Zhang, Liu, Wang, Lei, Wang, and Lu}{Zhang
  et~al\mbox{.}}{2020b}]%
        {zhang2020non}
\bibfield{author}{\bibinfo{person}{Pingping Zhang}, \bibinfo{person}{Wei Liu},
  \bibinfo{person}{Dong Wang}, \bibinfo{person}{Yinjie Lei},
  \bibinfo{person}{Hongyu Wang}, {and} \bibinfo{person}{Huchuan Lu}.}
  \bibinfo{year}{2020}\natexlab{b}.
\newblock \showarticletitle{{Non-rigid object tracking via deep multi-scale
  spatial-temporal discriminative saliency maps}}.
\newblock \bibinfo{journal}{\emph{Pattern Recognition}}  \bibinfo{volume}{100}
  (\bibinfo{year}{2020}), \bibinfo{pages}{107130}.
\newblock


\bibitem[\protect\citeauthoryear{Zhang, Liu, and Hu}{Zhang
  et~al\mbox{.}}{2021c}]%
        {zhang2021transfuse}
\bibfield{author}{\bibinfo{person}{Yundong Zhang}, \bibinfo{person}{Huiye Liu},
  {and} \bibinfo{person}{Qiang Hu}.} \bibinfo{year}{2021}\natexlab{c}.
\newblock \showarticletitle{{Transfuse: Fusing transformers and cnns for
  medical image segmentation}}.
\newblock \bibinfo{journal}{\emph{arXiv preprint arXiv:2102.08005}}
  (\bibinfo{year}{2021}).
\newblock


\bibitem[\protect\citeauthoryear{Zhang, Lin, Xu, Jin, Lu, and Fan}{Zhang
  et~al\mbox{.}}{2021b}]%
        {zhang2020bilateral}
\bibfield{author}{\bibinfo{person}{Zhao Zhang}, \bibinfo{person}{Zheng Lin},
  \bibinfo{person}{Jun Xu}, \bibinfo{person}{Wen-Da Jin},
  \bibinfo{person}{Shao-Ping Lu}, {and} \bibinfo{person}{Deng-Ping Fan}.}
  \bibinfo{year}{2021}\natexlab{b}.
\newblock \showarticletitle{Bilateral attention network for RGB-D salient
  object detection}.
\newblock \bibinfo{journal}{\emph{IEEE Transactions on Image Processing}}
  \bibinfo{volume}{30} (\bibinfo{year}{2021}), \bibinfo{pages}{1949--1961}.
\newblock


\bibitem[\protect\citeauthoryear{Zhao, Jiang, Jia, Torr, and Koltun}{Zhao
  et~al\mbox{.}}{2020a}]%
        {zhao2020point}
\bibfield{author}{\bibinfo{person}{Hengshuang Zhao}, \bibinfo{person}{Li
  Jiang}, \bibinfo{person}{Jiaya Jia}, \bibinfo{person}{Philip Torr}, {and}
  \bibinfo{person}{Vladlen Koltun}.} \bibinfo{year}{2020}\natexlab{a}.
\newblock \showarticletitle{Point transformer}.
\newblock \bibinfo{journal}{\emph{arXiv preprint arXiv:2012.09164}}
  (\bibinfo{year}{2020}).
\newblock


\bibitem[\protect\citeauthoryear{Zhao, Li, Liu, Bing, Chen, Snoek, and
  Tighe}{Zhao et~al\mbox{.}}{2021}]%
        {zhao2021tuber}
\bibfield{author}{\bibinfo{person}{Jiaojiao Zhao}, \bibinfo{person}{Xinyu Li},
  \bibinfo{person}{Chunhui Liu}, \bibinfo{person}{Shuai Bing},
  \bibinfo{person}{Hao Chen}, \bibinfo{person}{Cees~GM Snoek}, {and}
  \bibinfo{person}{Joseph Tighe}.} \bibinfo{year}{2021}\natexlab{}.
\newblock \showarticletitle{TubeR: Tube-Transformer for Action Detection}.
\newblock \bibinfo{journal}{\emph{arXiv preprint arXiv:2104.00969}}
  (\bibinfo{year}{2021}).
\newblock


\bibitem[\protect\citeauthoryear{Zhao, Zhao, Li, and Chen}{Zhao
  et~al\mbox{.}}{2020c}]%
        {zhao2020depth}
\bibfield{author}{\bibinfo{person}{Jiawei Zhao}, \bibinfo{person}{Yifan Zhao},
  \bibinfo{person}{Jia Li}, {and} \bibinfo{person}{Xiaowu Chen}.}
  \bibinfo{year}{2020}\natexlab{c}.
\newblock \showarticletitle{Is depth really necessary for salient object
  detection?}. In \bibinfo{booktitle}{\emph{Proceedings of the 28th ACM
  International Conference on Multimedia}}. \bibinfo{pages}{1745--1754}.
\newblock


\bibitem[\protect\citeauthoryear{Zhao, Cao, Fan, Cheng, Li, and Zhang}{Zhao
  et~al\mbox{.}}{2019}]%
        {zhao2019contrast}
\bibfield{author}{\bibinfo{person}{Jia-Xing Zhao}, \bibinfo{person}{Yang Cao},
  \bibinfo{person}{Deng-Ping Fan}, \bibinfo{person}{Ming-Ming Cheng},
  \bibinfo{person}{Xuan-Yi Li}, {and} \bibinfo{person}{Le Zhang}.}
  \bibinfo{year}{2019}\natexlab{}.
\newblock \showarticletitle{{Contrast prior and fluid pyramid integration for
  RGBD salient object detection}}. In \bibinfo{booktitle}{\emph{Proceedings of
  the IEEE Conference on Computer Vision and Pattern Recognition}}.
  \bibinfo{pages}{3927--3936}.
\newblock


\bibitem[\protect\citeauthoryear{Zhao, Zhang, Pang, Lu, and Zhang}{Zhao
  et~al\mbox{.}}{2020b}]%
        {zhao2020single}
\bibfield{author}{\bibinfo{person}{Xiaoqi Zhao}, \bibinfo{person}{Lihe Zhang},
  \bibinfo{person}{Youwei Pang}, \bibinfo{person}{Huchuan Lu}, {and}
  \bibinfo{person}{Lei Zhang}.} \bibinfo{year}{2020}\natexlab{b}.
\newblock \showarticletitle{{A single stream network for robust and real-time
  rgb-d salient object detection}}. In \bibinfo{booktitle}{\emph{European
  Conference on Computer Vision}}. Springer, \bibinfo{pages}{646--662}.
\newblock


\bibitem[\protect\citeauthoryear{Zheng, Cham, and Cai}{Zheng
  et~al\mbox{.}}{2021a}]%
        {zheng2021tfill}
\bibfield{author}{\bibinfo{person}{Chuanxia Zheng}, \bibinfo{person}{Tat-Jen
  Cham}, {and} \bibinfo{person}{Jianfei Cai}.}
  \bibinfo{year}{2021}\natexlab{a}.
\newblock \showarticletitle{TFill: Image Completion via a Transformer-Based
  Architecture}.
\newblock \bibinfo{journal}{\emph{arXiv preprint arXiv:2104.00845}}
  (\bibinfo{year}{2021}).
\newblock


\bibitem[\protect\citeauthoryear{Zheng, Lu, Zhao, Zhu, Luo, Wang, Fu, Feng,
  Xiang, Torr, et~al\mbox{.}}{Zheng et~al\mbox{.}}{2021b}]%
        {zheng2021rethinking}
\bibfield{author}{\bibinfo{person}{Sixiao Zheng}, \bibinfo{person}{Jiachen Lu},
  \bibinfo{person}{Hengshuang Zhao}, \bibinfo{person}{Xiatian Zhu},
  \bibinfo{person}{Zekun Luo}, \bibinfo{person}{Yabiao Wang},
  \bibinfo{person}{Yanwei Fu}, \bibinfo{person}{Jianfeng Feng},
  \bibinfo{person}{Tao Xiang}, \bibinfo{person}{Philip~HS Torr},
  {et~al\mbox{.}}} \bibinfo{year}{2021}\natexlab{b}.
\newblock \showarticletitle{Rethinking semantic segmentation from a
  sequence-to-sequence perspective with transformers}. In
  \bibinfo{booktitle}{\emph{Proceedings of the IEEE/CVF Conference on Computer
  Vision and Pattern Recognition}}. \bibinfo{pages}{6881--6890}.
\newblock


\bibitem[\protect\citeauthoryear{Zhu, Cai, Huang, Li, and Li}{Zhu
  et~al\mbox{.}}{2019}]%
        {zhu2019pdnet}
\bibfield{author}{\bibinfo{person}{Chunbiao Zhu}, \bibinfo{person}{Xing Cai},
  \bibinfo{person}{Kan Huang}, \bibinfo{person}{Thomas~H Li}, {and}
  \bibinfo{person}{Ge Li}.} \bibinfo{year}{2019}\natexlab{}.
\newblock \showarticletitle{{PDNet: Prior-model guided depth-enhanced network
  for salient object detection}}. In \bibinfo{booktitle}{\emph{2019 IEEE
  International Conference on Multimedia and Expo (ICME)}}. IEEE,
  \bibinfo{pages}{199--204}.
\newblock


\bibitem[\protect\citeauthoryear{Zhu, Guo, Zhang, Wang, Huang, Qiao, Liu, Wang,
  and Tang}{Zhu et~al\mbox{.}}{2021}]%
        {zhu2021aaformer}
\bibfield{author}{\bibinfo{person}{Kuan Zhu}, \bibinfo{person}{Haiyun Guo},
  \bibinfo{person}{Shiliang Zhang}, \bibinfo{person}{Yaowei Wang},
  \bibinfo{person}{Gaopan Huang}, \bibinfo{person}{Honglin Qiao},
  \bibinfo{person}{Jing Liu}, \bibinfo{person}{Jinqiao Wang}, {and}
  \bibinfo{person}{Ming Tang}.} \bibinfo{year}{2021}\natexlab{}.
\newblock \showarticletitle{AAformer: Auto-Aligned Transformer for Person
  Re-Identification}.
\newblock \bibinfo{journal}{\emph{arXiv preprint arXiv:2104.00921}}
  (\bibinfo{year}{2021}).
\newblock


\end{thebibliography}
\end{document}